\documentclass{article}

\PassOptionsToPackage{numbers, compress}{natbib}

\usepackage[main, final]{neurips_2026}

\usepackage[utf8]{inputenc} 
\usepackage{xurl}
\usepackage{hyperref}       
\usepackage{url}            
\usepackage{booktabs}       
\usepackage{amsfonts}       
\usepackage{nicefrac}       
\usepackage{microtype}      
\usepackage{xcolor}         

\usepackage[T1]{fontenc}    

\usepackage{natbib}
\usepackage{enumitem}
\usepackage{tikz}
\usepackage{multirow}
\usetikzlibrary{arrows.meta,positioning,fit,calc,backgrounds,shapes.misc,matrix}
\usepackage{array}
\usepackage{tcolorbox}
\usepackage{adjustbox}
\tcbuselibrary{listingsutf8}
 \usepackage{amsmath}

\usepackage{wrapfig}  

\title{Improving Consistency in Retrieval Augmented Systems With Group Similarity Rewards}
\addtocontents{toc}{\protect\setcounter{tocdepth}{-1}}

\author{%
  Faisal Hamman$^{1,2}$ \quad
  Chenyang Zhu$^{1}$ \quad
  Anoop Kumar$^{1}$ \quad
  Xujun Peng$^{1}$ \\
  \textbf{Sanghamitra Dutta}$^{2}$ \quad
  \textbf{Daben Liu}$^{1}$ \quad
  \textbf{Alfy Samuel}$^{1}$ \\[4pt]
  $^{1}$Capital One \quad
  $^{2}$University of Maryland, College Park
}

\begin{document}

\maketitle

\begin{abstract}
Retrieval-Augmented Generation (RAG) systems are increasingly deployed in high-stakes domains where users expect outputs to be consistent across semantically equivalent queries. However, existing systems often exhibit significant inconsistencies due to variability in both the retriever and generator, undermining trust and reliability. In this work, we focus on \emph{information consistency}, i.e., the requirement that generated outputs convey the same core content across semantically equivalent inputs. We introduce a principled evaluation framework that decomposes RAG consistency into retriever-level, generator-level, and end-to-end components, enabling systematic identification of inconsistency sources.  To improve consistency, we propose  \textbf{P}araphrased \textbf{S}et Group Relative Policy Optimization (PS-GRPO), an RL approach that leverages multiple rollouts across paraphrased set to assign \emph{group similarity rewards}.   We leverage PS-GRPO to achieve Information
\textbf{Con}sistent \textbf{RAG} (Con-RAG), training the generator to produce consistent outputs across paraphrased queries and remain robust to retrieval-induced variability. Because exact reward computation over paraphrase sets is computationally expensive, we also introduce a scalable approximation method that retains effectiveness while enabling efficient, large-scale training. Empirical evaluations across  short-form, multi-hop, and long-form QA benchmarks demonstrate that Con-RAG significantly improves consistency while preserving accuracy compared to baselines, even in the absence of explicit ground-truth supervision. Our work provides practical solutions for evaluating and building reliable RAG systems for safety-critical deployments. 
\end{abstract}

\section{Introduction}
LLMs are increasingly used in open-domain applications where users expect predictable behavior and consistent outputs for semantically equivalent or paraphrased inputs. However, they frequently generate divergent responses to such variations, raising concerns about their reliability and robustness~\citep{novikova2025consistency,elazar-etal-2021-measuring,raj2025improving}.
RAG systems are particularly prone to such inconsistencies~\cite{perccin2025investigating}. RAG architectures combine a retriever and a generator: the retriever selects top-$k$ documents from a large corpus based on the query, and the generator synthesizes a response conditioned on those documents~\cite{gao2023retrieval}. Semantically similar queries can lead to different retrieved document sets or rankings, resulting in divergent outputs~\cite{perccin2025investigating}. \citet{weller2025theoretical} also highlights a theoretical bottleneck in embedding-based retrieval, showing that the expressivity of top-$k$ retrieval is fundamentally limited; underscoring the need for LLMs that are robust to retrieval inconsistencies.
Furthermore, even when the evidence is fixed, the generator may still produce inconsistent responses due to the non-deterministic nature and phrasing sensitivity of LLMs~\cite{razavi2025benchmarking}.

This inconsistency is particularly problematic in high-stakes domains where RAG systems are commonly deployed, e.g., healthcare, finance, legal~\cite{kim2025fostering}. Inconsistent outputs can erode trust, introduce liability risks, or even mislead users~\citep{kim2025fostering, novikova2025consistency}. For instance, a customer service RAG assistant may offer different instructions for “\textit{How do I close my savings account?}” and “\textit{What steps should I take to shut down my savings account?}” despite these queries being semantically equivalent~\cite{razavi2025benchmarking}.

\begin{figure}
    \centering   \includegraphics[height=3.8cm]{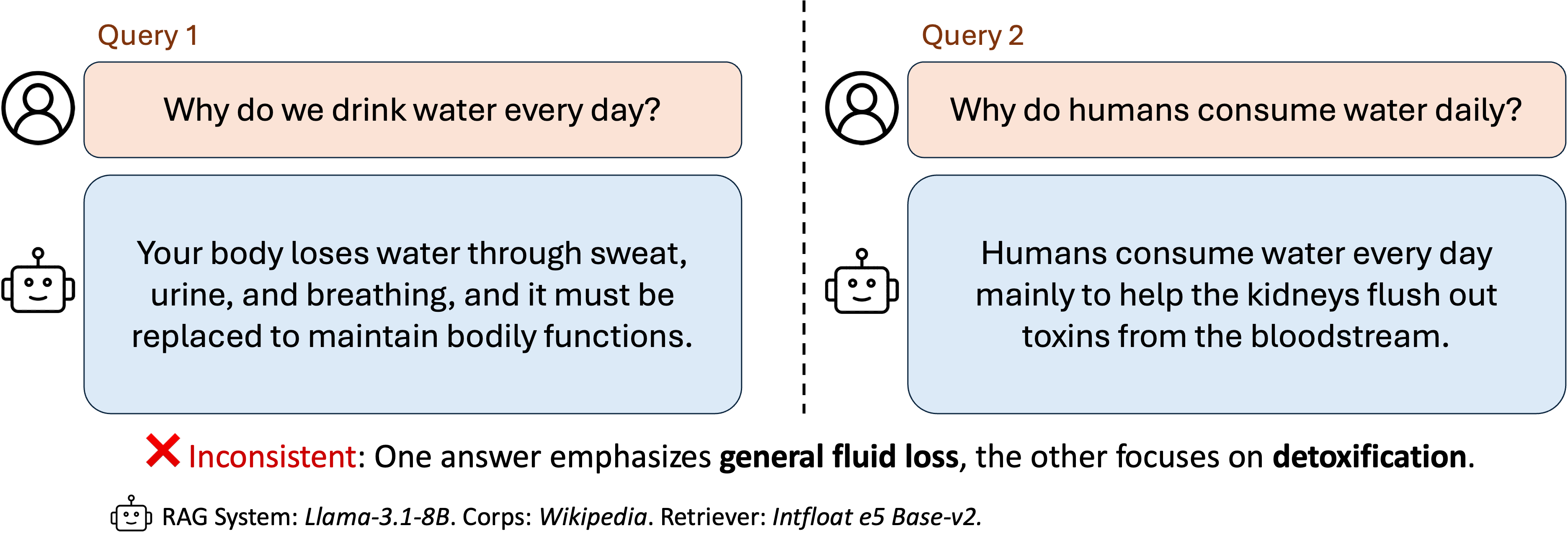}
    \caption{\small \textbf{Motivation.} Semantically equivalent queries lead to different outputs from a RAG system, despite both responses being factually correct. Such variation may be acceptable in many applications, but in certain high-stakes domains (e.g., healthcare, finance, legal) information consistency across equivalent inputs may be required to ensure reliability, user trust, and compliance.}
    \label{fig:motivation}
\end{figure}

In this work, we focus on \textit{information consistency}, i.e., the requirement that outputs convey the same core content and information across paraphrased inputs (see motivational Figure \ref{fig:motivation}). The relationship between consistency and accuracy varies across QA tasks. In short-form QA, where there is typically a single concise correct answer, accuracy and consistency are closely linked: a model that answers every paraphrase correctly is necessarily consistent. In contrast, for long-form QA tasks, where multiple valid answers may exist, consistency and accuracy can become orthogonal dimensions: a model can be accurate yet inconsistent, giving different but valid answers across paraphrases, or consistent yet unhelpful, repeating the same wrong or generic answer. Hence, in open-ended tasks, enforcing information consistency becomes a key desideratum alongside answer quality.

\begin{figure*}
    \centering   
    \includegraphics[width=0.95\textwidth]{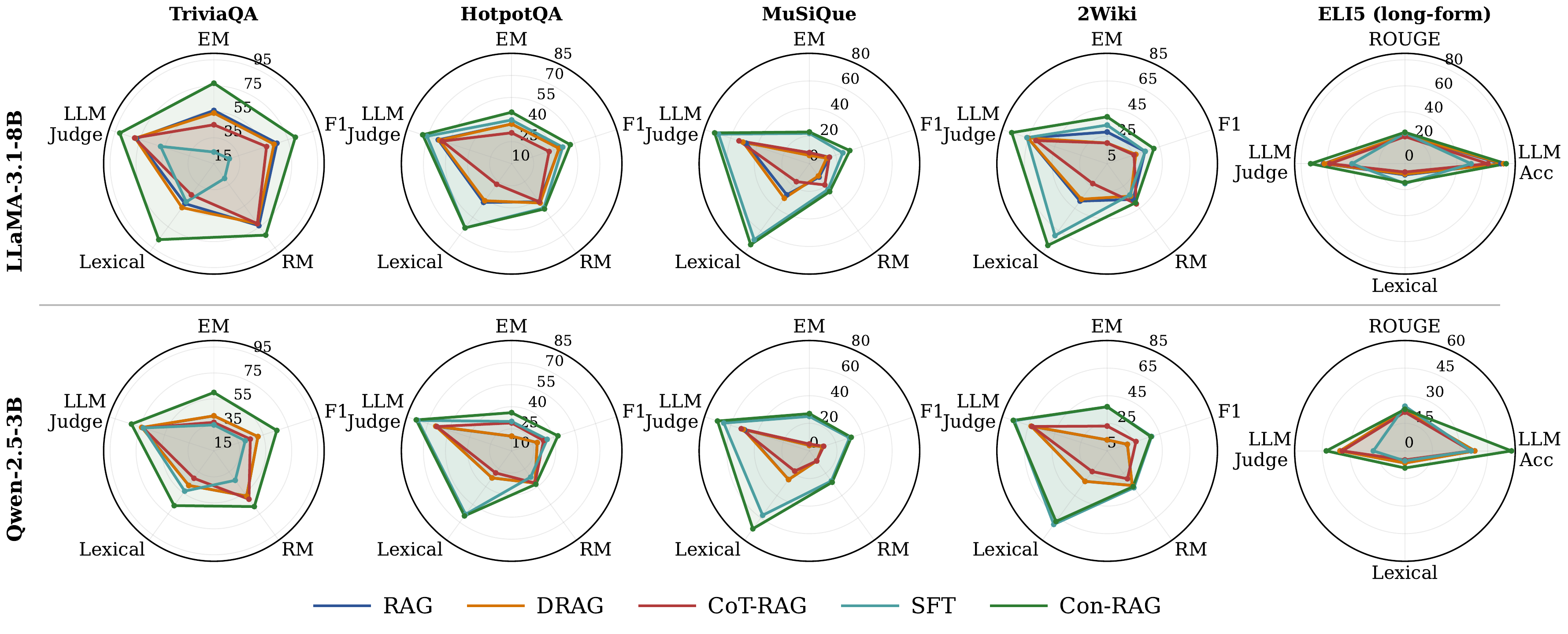}
    \caption{\small \textbf{Comparison between Con-RAG and baselines across accuracy and consistency dimensions on LLaMA-3.1-8B and Qwen-2.5-3B.} Each plot summarizes performance on a single dataset using accuracy measures (Exact Match, Token F1, Relaxed Match) and end-to-end information consistency (measured lexically (BLEU) and LLM-as-a-judge). Con-RAG outperforms prior methods across models, achieving higher factual accuracy and more consistent responses across paraphrased inputs (see Table~\ref{tab:main-results} \&~\ref{tab:main-results-longform} for full results including long form Natural Questions dataset).}
    \label{fig:raderplot}
\end{figure*}

Given the practical importance of consistent outputs, we aim to address the following question: \textit{How can we measure and improve the information consistency of RAG system outputs across semantically equivalent inputs, without compromising factual accuracy?} To tackle this, we introduce an evaluation framework that decomposes consistency into retriever-level and generator-level components, and propose a reinforcement learning approach to optimize for consistency using group similarity rewards.  Our contributions can be summarized as follows:

\begin{itemize}[leftmargin=*, topsep=0pt,itemsep=0pt]
    \item \textbf{A Framework for Measuring Consistency in RAG Systems.} We present a principled framework to evaluate consistency in RAG systems by disentangling three components: retriever consistency (Jaccard overlap of documents), generator consistency (LLM outputs given fixed context), and end-to-end consistency. We instantiate it with lexical (e.g., BLEU) and LLM-as-a-judge similarity, which lets us attribute inconsistency to the retriever or the generator, quantify how retrieval variation propagates to the final output (see Section~\ref{sec:measure}).

\item \textbf{Con-RAG: Improving Consistency via Paraphrased Set GRPO.}
To improve consistency across semantically equivalent queries, we propose \textbf{P}araphrased \textbf{S}et GRPO, an RL approach that leverages multiple rollouts across a set of paraphrased inputs to assign \emph{group similarity rewards}. This forms the core of our Information \textbf{Con}sistent \textbf{RAG} (Con-RAG) framework (see Figure~\ref{fig:system-model}). Due to complexity of computing the rewards, we introduce a relaxed approximation by subsampling paraphrases and rollouts, reducing the number of comparisons from quadratic to linear in the number of paraphrases. This allows us to train Con-RAG efficiently on large datasets while preserving reward fidelity (see Section~\ref{sec:method}).
\item \textbf{Empirical Evaluation.} We conduct an extensive evaluation of Con-RAG across six QA benchmarks: short-form, multi-hop, and long-form QA tasks on Llama-3.1 and Qwen-2.5 model families\footnote{Code: \url{https://github.com/CapitalOne-Research/con-rag}} (see Figure~\ref{fig:raderplot}).  Our results show that Con-RAG significantly improves both end-to-end and generator consistency compared to baselines,  while improving or preserving accuracy. This also holds for long-form QA tasks trained without ground-truth supervision.

\end{itemize}

\subsection{Related Work} 
\textbf{Consistency in Language Models.} Consistency has emerged as a key concern for safety and reliability in high-stakes LLM deployment~\cite{kim2025fostering,novikova2025consistency}. Prior work has introduced various notions of consistency. Logical consistency refers to the ability of the model to make decisions without logical contradiction~\cite{jang2022becel, li2019logic,asai2020logic,mitchell2022enhancing}. Factual consistency, often discussed as faithfulness or hallucination, considers whether model outputs contradict the source content~\cite{jang2022becel, wang2020asking, maynez2020faithfulness,tam2022evaluating}. Self-consistency evaluates whether similar inputs yield stable explanations~\cite{parcalabescu2023measuring}. Nonlogical forms of consistency, such as moral consistency, assess coherence of values across contexts~\cite{bonagiri2024sage, arvanitis2020consistency}. Prediction consistency examines whether a model’s predictions remain stable across multiple fine-tuned variants that achieve similar overall performance~\cite{hamman2025quantifying,gomez2024algorithmic}. Closest to our work is semantic consistency, which measures output stability under semantically equivalent inputs like paraphrases. This has been evaluated using datasets like ParaRel~\citep{elazar-etal-2021-measuring} and metrics such as BERTScore, entailment scores, and LLM judges~\citep{raj2022measuring, rabinovich2023predicting, kuhn2023semantic}. Approaches to improve semantic consistency include custom losses~\citep{elazar-etal-2021-measuring}, knowledge distillation from consistent teachers~\citep{raj2025improving}, and synthetic data supervision~\citep{zhao2024improving}. We refer to a recent survey exploring current landscape and future directions in consistency in LLMs~\cite{novikova2025consistency}.

\textbf{Consistency in RAG Systems.} RAG improves factual accuracy by conditioning outputs on retrieved evidence~\citep{guu2020retrieval, karpukhin2020dense, lewis2020retrieval}. However, it introduces new sources of inconsistency due to retriever sensitivity and generator (LLM) variability. Despite growing use in high-stakes applications, information consistency in RAG remains underexplored, with the exception of a few notable studies addressing robustness in retrieval or prompt-level variation~\citep{hsiaragged, zhang2025qe, hu2024prompt, perccin2025investigating}. 
Orthogonal work improves RAG accuracy, rather than consistency,  through retrieval-aware training or adaptive retrieval~\citep{lin2024radit,li2025ragddr,asai2024self,jin2025searchr,wang2025speculative}. Our work aims to evaluate and improve information consistency in RAG, leveraging an RL-based optimization with group similarity rewards. Our approach builds on recent advances in RL for LLMs~\citep{kaufmann2024survey}, particularly GRPO~\citep{shao2024deepseekmath}, which trains on verifiable reward assignment across outputs. Our framework improves information consistency across semantically equivalent inputs without requiring ground-truth labels, unlike prior consistency-improvement methods~\citep{raj2025improving,zhao2024improving}. Recent label-free post-training methods also learn from model-derived signals, such as self-certainty, majority-vote pseudo-labels, self-generated targets, or sharpened base distributions~\citep{zhao2026learning,zuo2026ttrl,li2025model,pmlr-v306-chen26k}; these rewards are defined on a single input, whereas PS-GRPO's reward is relational across paraphrased inputs.

\section{Main Contributions}
In this section, we first define a framework to measure consistency in RAG systems by isolating retriever, generator, and end-to-end contributions (see Section~\ref{sec:measure}), then introduce our Con-RAG method to improve consistency via group similarity rewards and its relaxation (see Section~\ref{sec:method}).
\subsection{Measuring Consistency in RAG
Systems}\label{sec:measure}
We consider a RAG system composed of a retriever $ R $ and a generator (LLM). Given a user query $ q $, the system first retrieves a set of top-$ k $ documents from a corpus $ \mathcal{D} $, and then generates an output $ y = \text{LLM}(q, R(q))$ conditioned on these documents:$ R(q) = \{ d_1, \dots, d_k \} \subset \mathcal{D}.$ Let $ q_0 $ be a canonical input query, and let $ \mathcal{P}(q_0) = \{ p_1, p_2, \dots, p_n \} $ denote a set of paraphrased or semantically equivalent inputs. Our goal is to assess the \emph{output consistency} of the RAG system across this paraphrased set.

\textbf{Retriever Consistency.}
Retriever consistency measures how stable retrieval is across paraphrased queries. Let $R(p_i)$ be the set of documents retrieved for $p_i$; for each pair $(p_i, p_j)$, we compare $R(p_i)$ and $R(p_j)$ using \emph{Jaccard similarity}~\cite{gower1986metric}, the ratio of the size of their intersection to the size of their union. Jaccard similarity equals $1$ when both paraphrases retrieve exactly the same documents and $0$ when they share none, and normalizing by the union makes it comparable across retrieval depths. We define retriever consistency as the average Jaccard similarity over all unique paraphrase pairs:
\[
\mathcal{C}_{\text{ret}}(q_0)= \frac{2}{n(n-1)} \sum_{i<j} \frac{|R(p_i) \cap R(p_j)|}{|R(p_i) \cup R(p_j)|}.
\]

\textbf{End-to-End RAG Consistency.}
Let $y_i = \text{LLM}(p_i, R(p_i))$ denote the output of the RAG system for paraphrase $p_i$. End-to-end consistency measures alignment across outputs when the entire pipeline is allowed to vary, each paraphrase $p_i$ is passed to the retriever, which may return a different document set $R(p_i)$, and the generator then conditions on this evidence to produce $y_i$. Formally, we compute pairwise similarity across all outputs: $$\mathcal{C}_{\text{e2e}}(q_0) = \frac{1}{n(n-1)} \sum_{i \neq j} \text{sim}(y_i, y_j).$$

This captures the overall stability of the RAG system under paraphrased inputs, reflecting the combined variability introduced by both retrieval and generation. The similarity function $\mathrm{sim}(\cdot,\cdot)$ can be instantiated with lexical metrics (e.g., BLEU, ROUGE), embedding-based similarity (e.g., BERTScore), entailment scores from NLI models, or an LLM-as-a-judge that assesses whether $y_i$ and $y_j$ convey the same information.

\textbf{Generator (LLM) Consistency.}
To isolate the generator’s contribution, we can fix the retrieved documents across all paraphrases and measure similarity among the outputs, i.e., \( y_i^{\text{fixed}} = \text{LLM}(p_i, R(q_0)) \), and compute consistency over \( \{ y_1^{\text{fixed}}, \dots, y_n^{\text{fixed}} \} \). This captures how consistently the LLM alone responds to semantically equivalent inputs when conditioned on identical evidence. Conceptually, this is closely related to prior work on consistency in standalone LLMs, where the focus is on ensuring paraphrase-invariant outputs under similar prompts~\cite{elazar-etal-2021-measuring,raj2025improving, novikova2025consistency, razavi2025benchmarking}.

\subsection{Improving Consistency via Paraphrased Set GRPO}\label{sec:method}

Given a RAG system comprising a retriever \(R\) and generator (LLM), and a canonical query \(q_0\) with paraphrases \(\mathcal{P}(q_0)=\{p_1,\dots,p_n\}\), our goal is to maximize output consistency while preserving, or improving factual accuracy. We propose Paraphrased Set GRPO, an RL algorithm that leverages GRPO's multiple rollouts across paraphrased inputs to assign group similarity rewards. Our objective is to directly train the generator so that outputs across semantically equivalent inputs are consistent.

\textbf{Group Relative Policy Optimization.}  GRPO is an RL optimization algorithm that estimates advantage through group-normalized rewards rather than using a critic model~\cite{shao2024deepseekmath}. For a given query $q$, GRPO samples a group of $g$ rollouts, i.e., multiple possible completions generated from the policy under stochastic decoding (such as temperature or nucleus sampling), denoted by $\{o_1,\dots,o_g\}$, where each rollout is drawn as $o_i \sim \pi_\theta(\cdot \mid q)$. Each rollout receives a verifiable scalar reward $r_i = \text{Reward}(o_i \mid q)$, and the normalized advantage for each rollout is computed as $\hat{A}_i = (r_i - \mu_q)/\sigma_q$, where $\mu_q$ and $\sigma_q$ are the mean and standard deviation of rewards within the group. Let  $y_{i,1:|o_i|}$ denote the tokens of response $o_i$, and let $\rho_{i,t} = \pi_\theta(y_{i,t} \mid q, y_{i,<t}) \,/\, \pi_{\theta_{\text{old}}}(y_{i,t} \mid q, y_{i,<t})$ be the token-level importance ratio. The policy is then  optimized by maximizing the objective using these group-relative advantages, with an optional KL penalty to penalize deviation from the reference policy:
\begin{align}
\mathcal{L}(\theta)
= \frac{1}{g}\sum_{i=1}^g \sum_{t=1}^{|o_i|}
& \min\!\Big( \rho_{i,t}\hat{A}_i,\,
\text{clip}(\rho_{i,t},\,1-\epsilon,\,1+\epsilon)\hat{A}_i \Big)\nonumber  - \beta\,\mathbb{D}_{\mathrm{KL}}\!\big(\pi_\theta(\cdot\mid q)\,\big\|\,\pi_{\text{ref}}(\cdot\mid q)\big)\end{align}

\begin{figure*}[t]
    \centering   \includegraphics[height=5.3cm]{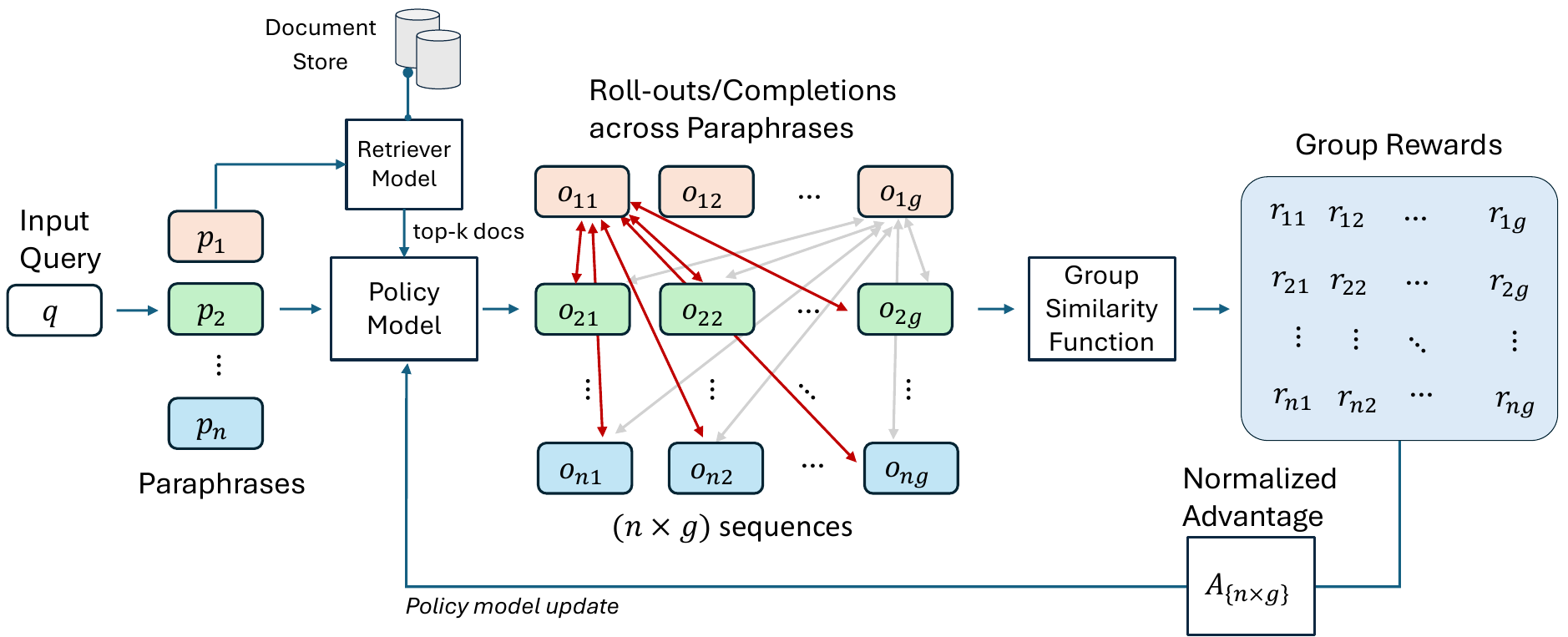}
    \caption{\small Overview of PS-GRPO and Information \textbf{Con}sistent \textbf{RAG} (Con-RAG) framework. A canonical query $q$ is expanded into a set of paraphrases $\{p_1,\dots,p_n\}$, each of which is passed through the policy model to generate $g$ sampled rollouts. For every rollout $o_{ij}$, we compute a group similarity reward $r_{ij}$ by averaging its similarity with outputs from other paraphrases of the same query (this produces an $n \times g$ reward matrix). Normalized advantages are then computed within each paraphrase set, and the policy model is updated.}
    \label{fig:system-model}
\end{figure*}

\textbf{PS-GRPO with Group Similarity Rewards.} PS-GRPO introduces a group-level objective that promotes consistent generation across semantically equivalent queries.
It leverages the unique property of GRPO which generates \emph{extensive
rollouts per query}. We aggregate all rollouts from all paraphrases into a
group and compute similarity-based rewards \emph{across the paraphrase set}, so each output is rewarded according to its similarity with outputs generated for the
other paraphrases of the same canonical query.  For each canonical query $q_0$ with paraphrases $\mathcal{P}(q_0)=\{p_1,\dots,p_n\}$, the policy LLM $\pi_\theta$ generates $g$ rollouts per paraphrase: $o_{ij}\sim \pi_\theta(\cdot \mid p_i, R(p_i)), i\in\{1,\dots,n\},\ j\in\{1,\dots,g\}.$
Collect these into an $n\times g$ matrix $\{o_{ij}\}$ (total $n \times g$ rollouts).
We assign each rollout $o_{ij}$ a group similarity reward by averaging its similarity to all rollouts generated for the \emph{other} paraphrases (also see Figure~\ref{fig:system-model} for illustration): \begin{equation}
r_{ij}\;=\;\frac{1}{(n-1)g}\sum_{\substack{u=1\\u\neq i}}^{n}\ \sum_{m=1}^{g}\ \mathrm{sim}\!\big(o_{ij},\,o_{um}\big),
\end{equation} where $\mathrm{sim}(\cdot,\cdot)$ is the similarity function. In practice, we instantiate $\mathrm{sim}$ using the BLEU metric, motivated by recent findings that BLEU serves as a strong proxy for reward models in aligning LLMs with human preferences~\cite{chang2025bleuberi}. As further validated in our ablation study (see Table~\ref{tab:abl-metric}), lower-order BLEU outperformed alternative lexical similarity metrics, and matched BERTScore reward on information consistency at substantially lower compute cost (see Appendix~\ref{apx:bertscore}). Group-normalized advantages are then computed across each paraphrased rollout:$
\hat{A}_{ij}=(r_{ij}-\mu_i)/\sigma_i
$, with \(\mu_i, \sigma_i\) the mean and standard deviation of rewards for rollouts for $p_i$.
The policy is optimized with the standard GRPO clipped objective using \(\hat{A}_{ij}\)
and (optionally) a KL penalty to a reference policy with weight \(\beta\). If ground-truth answers are available (e.g., in short-form QA tasks), we extend the reward to improve consistency and accuracy. Specifically, for each rollout we define a combined weighted reward:
\begin{equation}\label{eqn:final-reward}
r_{ij}^{\text{final}} \;=\; \alpha\, r_{ij}^{\text{cons}} \;+\; \gamma\, \text{Acc}(o_{ij}, y^\star),    
\end{equation}
where $r_{ij}^{\text{cons}}$ is the group similarity reward, $y^\star$ is the ground-truth answer, and $\text{Acc}(\cdot,\cdot)$ is measured using token F1 score. Importantly, our method does not require ground truths to improve consistency: the accuracy reward term can be omitted, as demonstrated in our long-form QA experiments (see Section~\ref{sec:exp}), where questions are open-ended and no single ground-truth answer exists.

\begin{table*}[t]
\centering
\caption{\small 
\textbf{Disentangling sources of inconsistency in RAG systems (LLaMA-3.1-8B)}.   Retriever consistency is low across datasets, suggesting that paraphrased queries often retrieve non-overlapping documents. This introduces context variability that is reflected in the end-to-end consistency scores. Fixing retrieval improves consistency, but variation remains, revealing the generator’s sensitivity to input phrasing even with identical evidence. We present accuracy values in Table~\ref{tab:accuracy-table-llama} (also see Table~\ref{tbl:consistency-table_qwen} for Qwen-2.5-3B).}
\footnotesize
\begin{tabular}{l|cc|cc|cc}
\toprule
\multirow{2}{*}{\textbf{Dataset}} 
& \multicolumn{2}{c}{\textbf{End-to-End Consistency}} 
& \multicolumn{2}{c}{\textbf{Generator (LLM) Consistency}} 
& \multicolumn{1}{c}{\textbf{Retriever Consistency}} \\
\cmidrule(lr){2-3}\cmidrule(lr){4-5}\cmidrule(lr){6-6}
& Lexical & LLM-Judge & Lexical & LLM-Judge & Jaccard Overlap \\
\midrule
TriviaQA   & 53.0  & 77.8 & 67.3 & 88.5 & 32.5 \\
HotpotQA  & 42.5 & 62.5 & 53.7 & 71.9 & 46.0 \\
2Wiki      & 38.5 & 65.5 & 48.4 & 76.4 & 52.4 \\
MuSiQue    & 27.9 & 48.2 & 44.4 & 69.7 & 36.6 \\
ElI5    & 8.6 & 62.8 & 15.1 & 74.2 & 27.1 \\
NQ    & 14.2 & 53.2  & 22.4 & 68.2 & 38.9 \\
\bottomrule
\end{tabular}
\label{tbl:consistency-table}
\end{table*}

\textbf{Efficient Computation of Group Similarity Rewards for Scalable Training.}
Computing group similarity rewards can be expensive, especially in a training environment where rewards must be computed at every gradient step. For each rollout $o_{ij}$, computing its reward requires comparing against all $(n-1)g$ rollouts from the other paraphrases. At the query level, with $n$ paraphrases and $g$ rollouts each, the naive total cost is $n g \times (n-1)g \;=\; n(n-1)g^2$ similarity computations. For example, with $n=5$ and $g=6$ amounts to $720$ comparisons for a single query. Exploiting symmetry reduces this to
$\tfrac{1}{2}\,n(n-1)g^2,$ but the cost still scales quadratically with $n$ and $g$.
To make training feasible, we introduce a \emph{relaxed group similarity reward}. Instead of averaging over all cross-paraphrase comparisons, for each rollout $o_{ij}$ we uniformly subsample $\kappa$ paraphrases $K {\subset} \{1,\dots,n\}\setminus\{i\}$ and $s$ rollouts per chosen paraphrase $S_u \subset \{1,\dots,g\}$, and approximate: $$\tilde r_{ij}\;=\;\frac{1}{\kappa s}\sum_{u\in K}\ \sum_{m\in S_u}\ \mathrm{sim}(o_{ij},\,o_{um}),$$  which is an unbiased estimator under uniform sampling. This reduces the per-query cost from $O(n(n-1)g^2)$ to $O(n g \kappa s)$, if $\kappa \ll n{-}1$ and $s \ll g$. In practice, this approximation preserves the training signal while keeping the reward computation tractable. We conduct ablations to study the impact of this subsampling on both performance and computational cost across a wide range of $(\kappa, s)$ settings (Appendix~\ref{apx:compute_ablation}).

\section{Experiments}\label{sec:exp}
In this section, we describe our experimental setup to evaluate Con-RAG on various QA tasks, describing datasets, paraphrase generation, consistency metrics, training details, and baselines.\footnote{Code: \url{https://github.com/CapitalOne-Research/con-rag}}

\textbf{Datasets.} We evaluate our approach across three types of question answering (QA) tasks: Short-form QA tasks: TriviaQA~\citep{triviaqa} and HotpotQA~\citep{yang2018hotpotqa}, both requiring concise fact-based answers. Multi-hop QA tasks: 2WikiMultiHopQA~\citep{ho2020constructing} and MuSiQue~\citep{trivedi2022musique}, which involve reasoning over multiple pieces of evidence. Long-form QA tasks: ELI5~\citep{fan2019eli5} and Natural Questions (NQ)~\cite{kwiatkowski2019natural}, where answers are open-ended and typically span multiple sentences. None of these datasets provide paraphrased versions of the input questions. Hence, we synthetically generate paraphrases for each query using LLaMA-3.1-70B. Specifically, for each original query $q_0$, we generate $n$ paraphrases $\mathcal{P}(q_0) = \{p_1, \dots, p_n\}$, enabling us to simulate semantically equivalent inputs (see Appendix~\ref{apx:paraphrase} for prompts and details).

\textbf{Setup.}
Our RAG system consists of a LLaMA-3.1-8B and Qwen-2.5-3B model serving as the generator, and a dense retriever built on top of the intfloat/e5-base-v2 embedding model~\citep{wang2022text}. We use KILT Wikipedia snapshot~\citep{fb_kilt} as our document corpus, where each article is segmented into chunks of 512 tokens. 
At inference time, the retriever selects the top-$k = 5$ documents per query, which are then appended to the prompt for generation. To isolate  effects from sampling inconsistencies, we use deterministic decoding for all evaluations.

\begin{table*}[t]
\centering
\caption{\small \textbf{Comparison between Con-RAG and Baselines} (Short-form QA, LLaMA-3.1-8B). Lexical consistency measured via BLEU score while information consistency measured using an LLM-as-a-judge. Con-RAG is trained with a group similarity reward plus an accuracy reward (no KL), and consistently yields higher end-to-end and generator-only consistency while also improving accuracy over original queries (see radar plot illustration in Figure~\ref{fig:raderplot}). Refer to Table~\ref{tab:main-results-qwen} for results on Qwen-2.5-3B model.}
\footnotesize
\begin{tabular}{ll|ccc|cc|cc}
\toprule
 & & \multicolumn{3}{c}{\textbf{Accuracy (\%)}} 
 & \multicolumn{2}{c}{\shortstack[c]{\textbf{End-to-End}\\\textbf{Consistency (\%)}}} 
 & \multicolumn{2}{c}{\shortstack[c]{\textbf{Generator (LLM)}\\\textbf{Consistency (\%)}}} \\
\cmidrule(lr){3-5} \cmidrule(lr){6-7} \cmidrule(lr){8-9}
\textbf{\centering Dataset} & \textbf{Method} 
& EM & F1 & RM 
& Lexical & Inform. 
& Lexical & Inform. \\
\midrule
\multirow{5}{*}{TriviaQA}
& RAG          & 56.0 & 66.1 & 74.0 & 53.0 & 77.8 & 67.3 & 88.5 \\
& DRAG         & 54.0 & 63.7 & 72.0 & 56.8 & 78.7 & 68.2 & 88.2 \\
& CoT-RAG      & 45.0 & 57.7 & 72.0 & 44.6 & 79.2 & 57.7 & 85.0 \\
& SFT          & 24.0 & 27.5 & 29.0 & 51.3 & 58.2 & 77.8 & 81.2 \\
& \textbf{Con-RAG } & \textbf{77.0} & \textbf{81.0} & \textbf{83.0} & \textbf{87.3} & \textbf{91.3} & \textbf{91.2} & \textbf{93.0} \\
\midrule
\multirow{5}{*}{HotpotQA}
& RAG          & 37.0 & 44.1 & 42.0 & 42.5 & 62.5 & 53.7 & 71.9 \\
& DRAG         & 37.0 & 43.8 & 43.0 & 41.1 & 61.6 & 50.5 & 73.1 \\
& CoT-RAG      & 31.0 & 36.8 & 42.0 & 27.3 & 59.6 & 36.1 & 68.9 \\
& SFT          & 39.7 & 46.5 & 47.2 & \textbf{63.9} & 70.5 & 72.2 & 78.5 \\
& \textbf{Con-RAG } & \textbf{45.0} & \textbf{51.9} & \textbf{48.0} & \textbf{63.9} & \textbf{73.6} & \textbf{80.9} & \textbf{88.2} \\
\midrule
\multirow{5}{*}{MuSiQue}
& RAG          &  8.0 & 15.3 & 12.0 & 27.9 & 48.2 & 44.4 & 69.7 \\
& DRAG         &  6.0 & 13.1 & 11.0 & 31.0 & 50.7 & 42.9 & 70.0 \\
& CoT-RAG      &  8.0 & 15.2 & 19.0 & 16.1 & 53.7 & 29.2 & 67.7 \\
& SFT          & 22.0 & 25.5 & 23.0 & 68.1 & 69.3 & 77.8 & 79.8 \\
& \textbf{Con-RAG } & \textbf{23.0} & \textbf{30.8} & \textbf{25.0} & \textbf{72.5} & \textbf{72.3} & \textbf{91.4} & \textbf{92.7} \\
\midrule
\multirow{5}{*}{2Wiki}
& RAG          & 28.0 & 33.9 & 37.0 & 38.5 & 65.5 & 48.4 & 76.4 \\
& DRAG         & 20.0 & 26.9 & 34.0 & 36.8 & 65.5 & 49.3 & 76.1 \\
& CoT-RAG      & 20.0 & 25.5 & 41.0 & 22.8 & 59.3 & 29.9 & 67.8 \\
& SFT          & 33.0 & 34.0 & 33.0 & 69.4 & 66.2 & 84.4 & 83.3 \\
& \textbf{Con-RAG } & \textbf{39.0} & \textbf{40.6} & \textbf{40.0} & \textbf{78.2} & \textbf{77.8} & \textbf{94.1} & \textbf{95.5} \\
\bottomrule
\end{tabular}
\label{tab:main-results}
\end{table*}

\textbf{Evaluating Consistency in RAG Systems.} We evaluate performance along two dimensions: accuracy and consistency. For short-form and multi-hop QA datasets, accuracy is measured using: (i) Exact Match (EM),
(ii) token F1 score, and (iii) Relaxed Match (RM), which considers an answer correct if the ground truth answer appears anywhere in the output. For long-form QA (e.g., ELI5, NQ), where answers are open-ended and may be phrased in diverse ways, EM/F1/RM are too restrictive. Instead, we evaluate accuracy using: (i) ROUGE, to capture content overlap with reference answers, and (ii) an LLM-as-a-judge, in which a strong model (LLaMA-3.3-70B) assesses whether the generated answer is factually correct; we report this as LLM-Acc. Consistency is evaluated at three levels (disentangling contributions from the retriever and generator):
(i) End-to-end consistency, where each paraphrase retrieves its own documents and we compute agreement between outputs (via BLEU for lexical consistency and an LLM-as-a-judge for information consistency; see Appendix~\ref{app:llm-judge} for prompts); (ii) Generator consistency, where retrieval is fixed across paraphrases and agreements across outputs are measured;
(iii) Retriever consistency, defined as the average Jaccard overlap between retrieved document sets across paraphrases (see Section~\ref{sec:measure}). We use paraphrase size $n = 5$ for evaluations.

We summarize consistency results across the datasets in Table~\ref{tbl:consistency-table}. We observe that the retriever consistency is relatively low across the datasets, indicating that paraphrases often retrieve non-overlapping sets of documents. This is reflected in the end-to-end consistency scores, which show that these small changes in query phrasing can result in different answers, due to shifts in both retrieved context and model generation.
To isolate the generator’s contribution, we also evaluate generator consistency under fixed retrieval (i.e., same documents across paraphrases). While consistency scores improve, substantial variability still remain, showing that even with identical evidence, the generator (LLM) exhibits sensitivity to input paraphrasing. We further quantify how retrieval variation propagates into end-to-end inconsistency through a per-query analysis in  Table~\ref{tab:jaccard-bins} in Appendix~\ref{apx:jaccard-bins}).
We report accuracy for original queries, paraphrased queries, and paraphrased queries with fixed documents in Table~\ref{tab:accuracy-table-llama}. Across these settings, accuracy remains relatively stable, with only minor fluctuations, suggesting that paraphrasing and retrieval shifts have limited impact on final answer correctness on average.

\begin{table*}[t]
\centering
\caption{\small \textbf{Comparison between Con-RAG and Baselines} (Long-form QA Task). Con-RAG is trained using only the group similarity rewards with a small KL regularizer (no accuracy supervision). Despite no ground-truth, it achieves the best information consistency and also maintains answer quality over baselines, whereas SFT on reference answers underperforms in this open-ended setting.}
\footnotesize
\begin{tabular}{ll|cc|cc|cc}
\toprule
 & & \multicolumn{2}{c}{\textbf{Accuracy (\%)}} 
 & \multicolumn{2}{c}{\shortstack[c]{\textbf{End-to-End}\\\textbf{Consistency (\%)}}} 
 & \multicolumn{2}{c}{\shortstack[c]{\textbf{Generator (LLM)}\\\textbf{Consistency (\%)}}} \\
\cmidrule(lr){3-4} \cmidrule(lr){5-6} \cmidrule(lr){7-8}
\textbf{Dataset} & \textbf{Method} 
& ROUGE & LLM-Acc 
& Lexical & Inform. 
& Lexical & Inform. \\
\midrule
\multirow{7}{*}{ELI5}
& RAG                 & 21.9 & 74.0 & 8.6  & 62.8 & 15.1 & 74.2 \\
& DRAG                & 22.0 & 76.0 & 8.0  & 62.2 & 15.0 & 72.5 \\
& CoT\text{-}RAG      & 20.9 & 64.0 & 6.4  & 57.8 & 10.3 & 71.0 \\
& SFT                 & 23.5 & 51.0 & \textbf{15.3} & 40.8 & 16.6 & 41.7 \\
& CoG                 & 23.1 & 55.3 &15.2  & 45.7 & 17.2 & 48.4  \\
& CAT                 & 22.8 & 58.1 & 14.7 & 49.0 & 16.3 & 50.9\\
& \textbf{Con\text{-}RAG} & \textbf{24.2} & \textbf{78.0} & 14.6 & \textbf{72.7} & \textbf{21.7} & \textbf{80.8} \\
\midrule
\multirow{7}{*}{NQ}
& RAG                 & 26.9 & \textbf{61.0 }& 14.2 & 53.2 & 22.4 & 68.2 \\
& DRAG                & 26.9 & 59.0 & 14.2 & 53.5 & 22.4 & 68.2 \\
& CoT\text{-}RAG      & 24.3 & 55.0 & 12.1 & 54.7 & 19.8 & 70.5 \\
& SFT                 & \textbf{36.8} & 22.0 & \textbf{37.1} & 36.2 & \textbf{56.2} & 55.0 \\
& CoG                 & 28.4 & 48.0 & 24.5 & 46.8 & 38.2 & 63.5 \\
& CAT                 & 27.1 & 53.0 & 20.8 & 52.1 & 33.5 & 66.8 \\
& \textbf{Con\text{-}RAG} 
                      & 23.5 & 59.0 & 22.8 & \textbf{59.3} & 32.3 & \textbf{75.5} \\
\bottomrule
\end{tabular}
\label{tab:main-results-longform}
\end{table*}

\textbf{Con-RAG Training Setup.} We train Con-RAG with BLEU as similarity function for computing group similarity rewards. For short-form and multi-hop QA tasks, we use unigram BLEU-1 and bigram BLEU-2 for long-form QA tasks to account for more contextual similarity across longer answers (see Appendix~\ref{apx:bleu} for background on BLEU-$n$). For short-form QA tasks, where ground-truth answers are available, we augment the similarity reward with an accuracy reward based on token F1 score, which we found to be more stable than other accuracy metrics. The final reward is computed using a weighted sum as defined in Eq.~\ref{eqn:final-reward}, with equal weights $(\alpha, \gamma = 1)$ for both consistency and accuracy. We set the KL regularization coefficient $\beta {=} 0$ for these tasks, following recent findings~\cite{hu2025open} suggesting that GRPO performs effectively without explicit KL penalties. In contrast, for long-form QA (ELI5, NQ), where questions are open-ended and multiple valid answers may exist, we exclude the accuracy reward and optimize solely for consistency using the group similarity rewards. To prevent reward hacking in the absence of ground-truth, we apply a small KL penalty with $\beta {=} 0.05$ to regularize the policy against a reference model (see Appendix~\ref{apx:kl} for a discussion and guidance on choosing $\beta$).

We use $n = 6$ paraphrases per canonical query and $g = 4$ rollouts per paraphrase. To make training scalable, we apply the relaxed approximation described in Section~\ref{sec:method} to estimate group similarity rewards. Specifically, we subsample $\kappa = 3$ paraphrases and $s = 1$ rollout per selected paraphrase when computing similarity, which significantly reduces the number of comparisons with minimal impact on reward quality.
We perform full model fine-tuning using the AdamW optimizer with a learning rate of \texttt{1e-6}. All training is conducted on LLaMA-3.1-8B and Qwen-2.5-3B. Full training configurations details are provided in Appendix~\ref{apx:training-config}.

\textbf{Baselines.}
All baselines share the same retriever, generator, and corpus.
\emph{Prompting:} (i) \textbf{RAG}, which appends the top-$k$ retrieved documents to the prompt; (ii) \textbf{DRAG} (Demonstrated RAG)~\cite{yue2024inference}, which adds few-shot demonstrations; (iii) \textbf{CoT-RAG} (Chain-of-Thought RAG)~\cite{zhao2024empirical}, which elicits intermediate reasoning before the answer.
\emph{Training-based:} (iv) \textbf{SFT} (Supervised Fine-Tuning)~\cite{chung2024scaling} on paraphrased queries paired with ground-truth answers (reference responses for long-form QA)
\emph{Consistency alignment:} (v) \textbf{CoG} (Chain of Guidance)~\cite{raj2025improving}, which fine-tunes the generator on paraphrase-consistent answers selected by an LLM ranking pipeline; (vi) \textbf{CAT} (Consistency Alignment Training)~\cite{zhao2024improving}, which follows paraphrase-augmented SFT with a self-rewarded ranking loss that prefers responses judged correct against the reference. Both were proposed for standalone LLMs; we adapt them to RAG by conditioning on retrieved documents. We evaluate CoG and CAT on long-form QA, where Con-RAG is trained without ground truth, making the comparison against these reference-supervised consistency methods most informative.

\textbf{Results and Analysis.} We present our results across short-form and long-form QA tasks in Figure~\ref{fig:raderplot} and Tables~\ref{tab:main-results}. To show that consistency improvements do not come at the cost of answer quality, we report accuracy metrics on the original queries, avoiding generic but consistent  outputs. Our results demonstrate the following key observations:

\begin{wraptable}{r}{0.55\textwidth}
\centering
\caption{\small
\textbf{Effect of reward similarity metric on Con-RAG} (ELI5, Qwen-2.5-3B).
Lower-order BLEU better preserves semantic adequacy across paraphrases,
while stricter metrics (higher-order BLEU, EM) penalize valid rephrasings.
BLEU-2 yields the best consistency and accuracy.
}
\footnotesize
\begin{tabular}{l|cc|cc|cc}
\toprule
\textbf{Metric}
& \multicolumn{2}{c}{\textbf{Acc}}
& \multicolumn{2}{c}{\textbf{E2E Cons}}
& \multicolumn{2}{c}{\textbf{Gen Cons}} \\
\cmidrule(lr){2-3} \cmidrule(lr){4-5} \cmidrule(lr){6-7}
& Rouge & LLM & Lex & Inf & Lex & Inf \\
\midrule
BLEU-1      & \textbf{22.6} & 54.0 & 6.9 & 38.2 & 14.8 & \textbf{69.8} \\
BLEU-2      & 22.5 & \textbf{58.0} & \textbf{9.2} & \textbf{42.0} & \textbf{17.8} & 67.5 \\
BLEU-3      & 22.4 & 49.0 & 6.7 & 36.3 & 14.8 & 66.0 \\
BLEU-4      & 22.2 & 50.0 & 6.4 & 36.2 & 14.2 & 66.5 \\
ROUGE-L     & 22.1 & 46.0 & 6.1 & 35.2 & 13.6 & 65.2 \\
Exact Match & 22.1 & 49.0 & 6.6 & 37.7 & 14.4 & 66.2 \\
\bottomrule
\end{tabular}
\label{tab:abl-metric}
\end{wraptable}
\textit{Con-RAG improves both consistency and accuracy in short-form QA.} Across all short and multi-hop datasets, Con-RAG achieves significant gains in both end-to-end and generator-only consistency. For instance, on TriviaQA, end-to-end consistency (lexical/information) improves from $53.0/77.8$ (RAG) to $87.3/91.3$, while generator consistency reaches $91.2/93.0$. Notably, improvements are not achieved at the expense of accuracy. Con-RAG also achieves the highest EM, F1, and RM scores across all datasets (see Table~\ref{tab:main-results}). This indicates that optimizing consistency can also enhance model robustness, likely due to the implicit data augmentation effect of training across paraphrased inputs (see Appendix~\ref{apx:accuracy}). DRAG and CoT-RAG provide only modest consistency improvements and fail to match Con-RAG.

\textit{In Long-form QA, Con-RAG improves consistency even without ground-truth supervision.} On ELI5, despite being trained without any explicit ground-truth, Con-RAG improves both consistency and accuracy over baselines. Compared to RAG, Con-RAG increases lexical and information consistency while also achieving higher ROUGE and LLM-judged accuracy (see Table~\ref{tab:main-results-longform}). For NQ dataset, consistency improves without taking a hit in accuracy. In contrast, SFT trained on reference answers performs poorly, especially in terms of LLM-judge accuracy, highlighting the limitations of rigid supervision in open-ended QA. 
We further show that these gains transfer out of distribution: Con-RAG trained on one long-form dataset and evaluated on the other preserves accuracy close to the RAG baseline while still improving consistency (see Appendix~\ref{apx:ood}).

\textbf{Ablation Studies.}\label{sec:ablations}
To analyze design choices in Con-RAG, we run focused ablations on a lighter generator, Qwen-2.5-3B for fast, controlled sweeps.

\textbf{(1)} \textit{Varying similarity function used in the group reward.} We replace BLEU in the group similarity reward with alternative choices and measure resulting consistency/accuracy. We consider: BLEU-$n$ ($n\!\in\!\{1,2,3,4\}$), ROUGE-L, Exact Match (results are summarized in Table~\ref{tab:abl-metric}).  Lower order BLEU yields the best consistency and accuracy. We further replace BLEU with BERTScore, a semantic similarity reward, which achieves comparable LLM-Judge consistency and accuracy with lower lexical consistency, but at higher computational cost (see Appendix~\ref{apx:bertscore}). 

\textbf{(2)} \textit{Subsampling $(\kappa, s)$ sensitivity and compute trade-offs.} We analyze the sensitivity of Con-RAG to the subsampling parameters $(\kappa, s)$ used in the relaxed group-similarity reward, showing stable performance across a wide range of settings with increasing compute costs at larger budgets. Detailed results and discussion are provided in Appendix~\ref{apx:compute_ablation}.

\textbf{(3)} \textit{Comparison with standard GRPO.}
We compare PS-GRPO against standard GRPO trained with a pointwise BLEU reward to ground-truth answers. PS-GRPO consistently achieves higher accuracy and consistency, despite using no ground-truth supervision.  The learning signal from mutual agreement across paraphrases is more effective than simply pushing each response toward a single reference. Full results and discussion are provided in
Appendix~\ref{apx:com_PSGRPO_GRPO}.

\textbf{(4)} \textit{Varying short-form accuracy reward metrics.} On short-form QA, we study the effect of different reward signals on accuracy and consistency by conducting ablations with: (i) consistency term only training, (ii) accuracy term only training, and (iii) joint training with consistency plus accuracy. For the accuracy component, we compare token F1 (ours), EM, and RM (see Table~\ref{tab:abl-acc}). Best performance is achieved when
combining consistency with the token F1 reward. 

\textbf{(5)} \textit{Robustness to decoding strategy and retrieval depth.}
We evaluate how inference-time stochasticity impacts consistency and accuracy by sweeping the decoding temperature $T \in \{0.0, 0.5, 1.0, 2.0\}$ (Table~\ref{tab:abl-temp}) and by evaluating Con-RAG under greedy, top-$p$, top-$k$, and temperature sampling. We also vary the number of retrieved documents ($k \in \{5, 10, 15\}$). Con-RAG maintains consistency gains over the RAG baseline across all decoding strategies and retrieval depths (see Appendix~\ref{apx:robustness_pipeline}).

\begin{table*}[t]
\centering
\caption{\small \textbf{Effect of Accuracy Reward Variant on Con-RAG} (TriviaQA, Qwen-2.5-3B). We compare consistency-only training, accuracy-only training, and joint training with consistency plus various accuracy metrics. The best performance is achieved when combining consistency with the token F1 reward, which yields the highest accuracy and consistency values.}
\footnotesize
\begin{tabular}{lcc|ccc|cc|cc}
\toprule
\textbf{Reward Variant} & $\alpha$ & $\gamma$ 
& \multicolumn{3}{c}{\textbf{Accuracy (\%)}} 
& \multicolumn{2}{c}{\textbf{End-to-End Cons. (\%)}} 
& \multicolumn{2}{c}{\textbf{Generator Cons. (\%)}} \\
\cmidrule(lr){4-6}\cmidrule(lr){7-8}\cmidrule(lr){9-10}
& & & EM & F1 & RM & Lexical & Inform. & Lexical & Inform. \\
\midrule
Consistency only     & 1.0 & 0.0 & 51.5 & 53.2 & 59.0 & 59.9 & 79.0 & 78.7 & 88.0 \\
Accuracy only (F1)   & 0.0 & 1.0 & 54.0 & 56.0 & 60.4 & 52.0 & 75.0 & 62.0 & 84.1 \\
Consistency + EM     & 1.0 & 1.0 & 56.2 & 63.5 & 65.0 & 61.5 & 80.2 & 76.0 & 88.4 \\
Consistency + RM     & 1.0 & 1.0 & 57.0 & 64.0 & 66.0 & 62.3 & 80.5 & 77.0 & 88.5 \\
Consistency + F1 & 1.0 & 1.0 & \textbf{60.0} & \textbf{66.0} & \textbf{68.0} & \textbf{67.1} & \textbf{81.8} & \textbf{80.5} & \textbf{89.5} \\
\bottomrule
\end{tabular}
\label{tab:abl-acc}
\end{table*}

\section{Discussion}\label{sec:disc} \textbf{BLEU metric as a proxy reward signal.} BLEU metrics alone do not capture semantic consistency. The metric serves as a lightweight proxy signal that nudges the model towards more consistent outputs across paraphrases. Our use of BLEU is also driven by scalability considerations; group-level reward computation must be performed hundreds of times per optimization step, making an efficient metric essential. Moreover, prior work~\cite{chang2025bleuberi} shows that lower-order BLEU correlates surprisingly well with human preference signals (also see ablations in Table~\ref{tab:abl-metric}).  Importantly, even under such minimal reward, Con-RAG improves not only lexical overlap but also LLM-judge consistency, which directly evaluates information-level agreement. Stronger semantic rewards can be substituted, but at increasing cost: replacing BLEU with BERTScore yields comparable information consistency and accuracy at substantially higher compute (Appendix~\ref{apx:bertscore}), while an LLM-based consistency critic would be prohibitively expensive at this scale. In this work, we demonstrate that (1) group similarity rewards even without ground truths is an effective framework for optimizing consistency, and (2) even with a simple and efficient similarity function, the method produces measurable gains in information consistency.

\textbf{Beyond Lexical Rewards for Information Consistency.}  An important direction for future work is to develop reward signals that directly optimize information consistency without relying on lexical similarity. While LLM-as-a-judge metrics are promising, they introduce a trade-off between weak and strong supervision~\cite{burns2023weak}. Our BERTScore experiment (Appendix~\ref{apx:bertscore}) shows that semantic rewards preserve information consistency while tolerating lexical diversity, but at a higher compute cost. Ideally, we seek lightweight, efficient signals that can guide models toward consistent information content, such as entailment-based rewards or cheaper embedding-based approaches.

\textbf{Retrieval-induced variability.}
Our approach is retriever-agnostic and can be applied on top of any retrieval system. We show that even when the retriever is perfectly consistent, i.e., when we freeze the retrieved documents across paraphrases, substantial inconsistency remains at the generator level. This motivates our focus on improving the generator to remain robust to retrieval-induced variability. This design choice becomes more important as LLMs move toward agentic systems that interleave reasoning with calls to external tools such as web search~\cite{yao2022react, schick2023toolformer, singh2025agentic}. In these settings, retrieval is often provided by third-party, API-based search services that the system builder cannot modify or train, so variability in what is retrieved must be absorbed by the model itself. Methods that train only the generator, like ours, therefore remain applicable even when the retrieval component is a black box. We also note that our approach is not limited to RAG: it applies broadly to standalone LLMs, where paraphrase sensitivity and generation inconsistency persist even without retrieval. We focus on RAG because the variability introduced by the retriever amplifies these inconsistencies and makes the problem more critical in practice. When the retriever is accessible, jointly training or regularizing the retriever and generator is a promising direction for future work~\cite{lewis2020retrieval, lin2024radit, li2025ragddr}. By introducing a principled way to measure consistency and a scalable method to improve it, we take a step towards more reliable and user-aligned RAG systems. We discuss broader impacts in Appendix~\ref{apx:impact}.

\textbf{\textit{Acknowledgment}.}
Part of this work was done during an internship at Capital One and was funded by Capital One. Additional support was provided by the Ann G. Wylie Dissertation Fellowship from the University of Maryland Graduate School.

\bibliography{rag}

\begin{thebibliography}{69}
\providecommand{\natexlab}[1]{#1}
\providecommand{\url}[1]{\texttt{#1}}
\expandafter\ifx\csname urlstyle\endcsname\relax
  \providecommand{\doi}[1]{doi: #1}\else
  \providecommand{\doi}{doi: \begingroup \urlstyle{rm}\Url}\fi

\bibitem[Novikova et~al.(2025)Novikova, Anderson, Blili-Hamelin, Rosati, and Majumdar]{novikova2025consistency}
Jekaterina Novikova, Carol Anderson, Borhane Blili-Hamelin, Domenic Rosati, and Subhabrata Majumdar.
\newblock Consistency in language models: Current landscape, challenges, and future directions.
\newblock \emph{arXiv preprint arXiv:2505.00268}, 2025.

\bibitem[Elazar et~al.(2021)Elazar, Kassner, Ravfogel, Ravichander, Hovy, Sch{\"u}tze, and Goldberg]{elazar-etal-2021-measuring}
Yanai Elazar, Nora Kassner, Shauli Ravfogel, Abhilasha Ravichander, Eduard Hovy, Hinrich Sch{\"u}tze, and Yoav Goldberg.
\newblock Measuring and improving consistency in pretrained language models.
\newblock \emph{Transactions of the Association for Computational Linguistics}, 9:\penalty0 1012--1031, 2021.
\newblock \doi{10.1162/tacl_a_00410}.
\newblock URL \url{https://aclanthology.org/2021.tacl-1.60/}.

\bibitem[Raj et~al.(2025)Raj, Gupta, Rosati, and Majumdar]{raj2025improving}
Harsh Raj, Vipul Gupta, Domenic Rosati, and Subhabrata Majumdar.
\newblock Improving consistency in large language models through chain of guidance.
\newblock \emph{arXiv preprint arXiv:2502.15924}, 2025.

\bibitem[Per{\c{c}}in et~al.(2025)Per{\c{c}}in, Su, Syed, Howard, Kuvshinov, Schwinn, and Scholl]{perccin2025investigating}
Sezen Per{\c{c}}in, Xin Su, Qutub~Sha Syed, Phillip Howard, Aleksei Kuvshinov, Leo Schwinn, and Kay-Ulrich Scholl.
\newblock Investigating the robustness of retrieval-augmented generation at the query level.
\newblock \emph{arXiv preprint arXiv:2507.06956}, 2025.

\bibitem[Gao et~al.(2023)Gao, Xiong, Gao, Jia, Pan, Bi, Dai, Sun, Wang, and Wang]{gao2023retrieval}
Yunfan Gao, Yun Xiong, Xinyu Gao, Kangxiang Jia, Jinliu Pan, Yuxi Bi, Yixin Dai, Jiawei Sun, Haofen Wang, and Haofen Wang.
\newblock Retrieval-augmented generation for large language models: A survey.
\newblock \emph{arXiv preprint arXiv:2312.10997}, 2\penalty0 (1), 2023.

\bibitem[Weller et~al.(2025)Weller, Boratko, Naim, and Lee]{weller2025theoretical}
Orion Weller, Michael Boratko, Iftekhar Naim, and Jinhyuk Lee.
\newblock On the theoretical limitations of embedding-based retrieval.
\newblock \emph{arXiv preprint arXiv:2508.21038}, 2025.

\bibitem[Razavi et~al.(2025)Razavi, Soltangheis, Arabzadeh, Salamat, Zihayat, and Bagheri]{razavi2025benchmarking}
Amirhossein Razavi, Mina Soltangheis, Negar Arabzadeh, Sara Salamat, Morteza Zihayat, and Ebrahim Bagheri.
\newblock Benchmarking prompt sensitivity in large language models.
\newblock In \emph{European Conference on Information Retrieval}, pages 303--313. Springer, 2025.

\bibitem[Kim et~al.(2025)Kim, Vaughan, Liao, Lombrozo, and Russakovsky]{kim2025fostering}
Sunnie~SY Kim, Jennifer~Wortman Vaughan, Q~Vera Liao, Tania Lombrozo, and Olga Russakovsky.
\newblock Fostering appropriate reliance on large language models: The role of explanations, sources, and inconsistencies.
\newblock In \emph{Proceedings of the 2025 CHI Conference on Human Factors in Computing Systems}, pages 1--19, 2025.

\bibitem[Jang et~al.(2022)Jang, Kwon, and Lukasiewicz]{jang2022becel}
Myeongjun Jang, Deuk~Sin Kwon, and Thomas Lukasiewicz.
\newblock Becel: Benchmark for consistency evaluation of language models.
\newblock In \emph{Proceedings of the 29th International Conference on Computational Linguistics}, pages 3680--3696, 2022.

\bibitem[Li et~al.(2019)Li, Gupta, Mehta, and Srikumar]{li2019logic}
Tao Li, Vivek Gupta, Maitrey Mehta, and Vivek Srikumar.
\newblock A logic-driven framework for consistency of neural models.
\newblock \emph{arXiv preprint arXiv:1909.00126}, 2019.

\bibitem[Asai and Hajishirzi(2020)]{asai2020logic}
Akari Asai and Hannaneh Hajishirzi.
\newblock Logic-guided data augmentation and regularization for consistent question answering.
\newblock \emph{arXiv preprint arXiv:2004.10157}, 2020.

\bibitem[Mitchell et~al.(2022)Mitchell, Noh, Li, Armstrong, Agarwal, Liu, Finn, and Manning]{mitchell2022enhancing}
Eric Mitchell, Joseph~J Noh, Siyan Li, William~S Armstrong, Ananth Agarwal, Patrick Liu, Chelsea Finn, and Christopher~D Manning.
\newblock Enhancing self-consistency and performance of pre-trained language models through natural language inference.
\newblock \emph{arXiv preprint arXiv:2211.11875}, 2022.

\bibitem[Wang et~al.(2020)Wang, Cho, and Lewis]{wang2020asking}
Alex Wang, Kyunghyun Cho, and Mike Lewis.
\newblock Asking and answering questions to evaluate the factual consistency of summaries.
\newblock \emph{arXiv preprint arXiv:2004.04228}, 2020.

\bibitem[Maynez et~al.(2020)Maynez, Narayan, Bohnet, and McDonald]{maynez2020faithfulness}
Joshua Maynez, Shashi Narayan, Bernd Bohnet, and Ryan McDonald.
\newblock On faithfulness and factuality in abstractive summarization.
\newblock \emph{arXiv preprint arXiv:2005.00661}, 2020.

\bibitem[Tam et~al.(2022)Tam, Mascarenhas, Zhang, Kwan, Bansal, and Raffel]{tam2022evaluating}
Derek Tam, Anisha Mascarenhas, Shiyue Zhang, Sarah Kwan, Mohit Bansal, and Colin Raffel.
\newblock Evaluating the factual consistency of large language models through news summarization.
\newblock \emph{arXiv preprint arXiv:2211.08412}, 2022.

\bibitem[Parcalabescu and Frank(2023)]{parcalabescu2023measuring}
Letitia Parcalabescu and Anette Frank.
\newblock On measuring faithfulness or self-consistency of natural language explanations.
\newblock \emph{arXiv preprint arXiv:2311.07466}, 2023.

\bibitem[Bonagiri et~al.(2024)Bonagiri, Vennam, Govil, Kumaraguru, and Gaur]{bonagiri2024sage}
Vamshi~Krishna Bonagiri, Sreeram Vennam, Priyanshul Govil, Ponnurangam Kumaraguru, and Manas Gaur.
\newblock Sage: Evaluating moral consistency in large language models.
\newblock \emph{arXiv preprint arXiv:2402.13709}, 2024.

\bibitem[Arvanitis and Kalliris(2020)]{arvanitis2020consistency}
Alexios Arvanitis and Konstantinos Kalliris.
\newblock Consistency and moral integrity: A self-determination theory perspective.
\newblock \emph{Journal of Moral Education}, 49\penalty0 (3):\penalty0 316--329, 2020.

\bibitem[Hamman et~al.(2025)Hamman, Dissanayake, Mishra, Lecue, and Dutta]{hamman2025quantifying}
Faisal Hamman, Pasan Dissanayake, Saumitra Mishra, Freddy Lecue, and Sanghamitra Dutta.
\newblock Quantifying prediction consistency under fine-tuning multiplicity in tabular {LLM}s.
\newblock In \emph{Forty-second International Conference on Machine Learning}, 2025.
\newblock URL \url{https://openreview.net/forum?id=AXJnqocQpm}.

\bibitem[Gomez et~al.(2024)Gomez, Machado, Paes, and Calmon]{gomez2024algorithmic}
Juan~Felipe Gomez, Caio Machado, Lucas~Monteiro Paes, and Flavio Calmon.
\newblock Algorithmic arbitrariness in content moderation.
\newblock In \emph{Proceedings of the 2024 ACM Conference on Fairness, Accountability, and Transparency}, pages 2234--2253, 2024.

\bibitem[Raj et~al.(2022)Raj, Rosati, and Majumdar]{raj2022measuring}
Harsh Raj, Domenic Rosati, and Subhabrata Majumdar.
\newblock Measuring reliability of large language models through semantic consistency.
\newblock \emph{arXiv preprint arXiv:2211.05853}, 2022.

\bibitem[Rabinovich et~al.(2023)Rabinovich, Ackerman, Raz, Farchi, and Anaby-Tavor]{rabinovich2023predicting}
Ella Rabinovich, Samuel Ackerman, Orna Raz, Eitan Farchi, and Ateret Anaby-Tavor.
\newblock Predicting question-answering performance of large language models through semantic consistency.
\newblock \emph{arXiv preprint arXiv:2311.01152}, 2023.

\bibitem[Kuhn et~al.(2023)Kuhn, Gal, and Farquhar]{kuhn2023semantic}
Lorenz Kuhn, Yarin Gal, and Sebastian Farquhar.
\newblock Semantic uncertainty: Linguistic invariances for uncertainty estimation in natural language generation.
\newblock \emph{arXiv preprint arXiv:2302.09664}, 2023.

\bibitem[Zhao et~al.(2024{\natexlab{a}})Zhao, Yan, Sun, Xing, Wang, Meng, Cheng, Ren, and Yin]{zhao2024improving}
Yukun Zhao, Lingyong Yan, Weiwei Sun, Guoliang Xing, Shuaiqiang Wang, Chong Meng, Zhicong Cheng, Zhaochun Ren, and Dawei Yin.
\newblock Improving the robustness of large language models via consistency alignment.
\newblock \emph{arXiv preprint arXiv:2403.14221}, 2024{\natexlab{a}}.

\bibitem[Guu et~al.(2020)Guu, Lee, Tung, Pasupat, and Chang]{guu2020retrieval}
Kelvin Guu, Kenton Lee, Zora Tung, Panupong Pasupat, and Mingwei Chang.
\newblock Retrieval augmented language model pre-training.
\newblock In \emph{International conference on machine learning}, pages 3929--3938. PMLR, 2020.

\bibitem[Karpukhin et~al.(2020)Karpukhin, Oguz, Min, Lewis, Wu, Edunov, Chen, and Yih]{karpukhin2020dense}
Vladimir Karpukhin, Barlas Oguz, Sewon Min, Patrick~SH Lewis, Ledell Wu, Sergey Edunov, Danqi Chen, and Wen-tau Yih.
\newblock Dense passage retrieval for open-domain question answering.
\newblock In \emph{EMNLP (1)}, pages 6769--6781, 2020.

\bibitem[Lewis et~al.(2020)Lewis, Perez, Piktus, Petroni, Karpukhin, Goyal, K{\"u}ttler, Lewis, Yih, Rockt{\"a}schel, et~al.]{lewis2020retrieval}
Patrick Lewis, Ethan Perez, Aleksandra Piktus, Fabio Petroni, Vladimir Karpukhin, Naman Goyal, Heinrich K{\"u}ttler, Mike Lewis, Wen-tau Yih, Tim Rockt{\"a}schel, et~al.
\newblock Retrieval-augmented generation for knowledge-intensive nlp tasks.
\newblock \emph{Advances in neural information processing systems}, 33:\penalty0 9459--9474, 2020.

\bibitem[Hsia et~al.()Hsia, Shaikh, Wang, and Neubig]{hsiaragged}
Jennifer Hsia, Afreen Shaikh, Zora~Zhiruo Wang, and Graham Neubig.
\newblock Ragged: Towards informed design of scalable and stable rag systems.
\newblock In \emph{Forty-second International Conference on Machine Learning}.

\bibitem[Zhang et~al.(2025)Zhang, Sun, Yu, Zang, Zheng, Song, Li, and Xu]{zhang2025qe}
Kepu Zhang, Zhongxiang Sun, Weijie Yu, Xiaoxue Zang, Kai Zheng, Yang Song, Han Li, and Jun Xu.
\newblock Qe-rag: A robust retrieval-augmented generation benchmark for query entry errors.
\newblock \emph{arXiv preprint arXiv:2504.04062}, 2025.

\bibitem[Hu et~al.(2024)Hu, Wang, Shu, Paik, and Zhu]{hu2024prompt}
Zhibo Hu, Chen Wang, Yanfeng Shu, Hye-Young Paik, and Liming Zhu.
\newblock Prompt perturbation in retrieval-augmented generation based large language models.
\newblock In \emph{Proceedings of the 30th ACM SIGKDD Conference on Knowledge Discovery and Data Mining}, pages 1119--1130, 2024.

\bibitem[Lin et~al.(2024)Lin, Chen, Chen, Shi, Lomeli, James, Rodriguez, Kahn, Szilvasy, Lewis, Zettlemoyer, and tau Yih]{lin2024radit}
Xi~Victoria Lin, Xilun Chen, Mingda Chen, Weijia Shi, Maria Lomeli, Richard James, Pedro Rodriguez, Jacob Kahn, Gergely Szilvasy, Mike Lewis, Luke Zettlemoyer, and Wen tau Yih.
\newblock {RA}-{DIT}: Retrieval-augmented dual instruction tuning.
\newblock In \emph{The Twelfth International Conference on Learning Representations}, 2024.
\newblock URL \url{https://openreview.net/forum?id=22OTbutug9}.

\bibitem[Li et~al.(2025{\natexlab{a}})Li, Mei, Liu, Yan, Wang, Yu, Zeng, Chen, Yu, Liu, Sun, and Xiong]{li2025ragddr}
Xinze Li, Sen Mei, Zhenghao Liu, Yukun Yan, Shuo Wang, Shi Yu, Zheni Zeng, Hao Chen, Ge~Yu, Zhiyuan Liu, Maosong Sun, and Chenyan Xiong.
\newblock {RAG}-{DDR}: Optimizing retrieval-augmented generation using differentiable data rewards.
\newblock In \emph{The Thirteenth International Conference on Learning Representations}, 2025{\natexlab{a}}.
\newblock URL \url{https://openreview.net/forum?id=Pnktu2PBXD}.

\bibitem[Asai et~al.(2024)Asai, Wu, Wang, Sil, and Hajishirzi]{asai2024self}
Akari Asai, Zeqiu Wu, Yizhong Wang, Avi Sil, and Hannaneh Hajishirzi.
\newblock Self-rag: Learning to retrieve, generate, and critique through self-reflection.
\newblock In \emph{International conference on learning representations}, volume 2024, pages 9112--9141, 2024.

\bibitem[Jin et~al.(2025)Jin, Zeng, Yue, Yoon, Arik, Wang, Zamani, and Han]{jin2025searchr}
Bowen Jin, Hansi Zeng, Zhenrui Yue, Jinsung Yoon, Sercan~O Arik, Dong Wang, Hamed Zamani, and Jiawei Han.
\newblock Search-r1: Training {LLM}s to reason and leverage search engines with reinforcement learning.
\newblock In \emph{Second Conference on Language Modeling}, 2025.
\newblock URL \url{https://openreview.net/forum?id=Rwhi91ideu}.

\bibitem[Wang et~al.(2025)Wang, Wang, Le, Zheng, Mishra, Perot, Zhang, Mattapalli, Taly, Shang, et~al.]{wang2025speculative}
Zilong~Ryan Wang, Zifeng Wang, Long Le, Huaixiu~Steven Zheng, Swaroop Mishra, Vincent Perot, Yuwei Zhang, Anush Mattapalli, Ankur Taly, Jingbo Shang, et~al.
\newblock Speculative rag: Enhancing retrieval augmented generation through drafting.
\newblock In \emph{International Conference on Learning Representations}, volume 2025, pages 18483--18505, 2025.

\bibitem[Kaufmann et~al.(2024)Kaufmann, Weng, Bengs, and H{\"u}llermeier]{kaufmann2024survey}
Timo Kaufmann, Paul Weng, Viktor Bengs, and Eyke H{\"u}llermeier.
\newblock A survey of reinforcement learning from human feedback.
\newblock 2024.

\bibitem[Shao et~al.(2024)Shao, Wang, Zhu, Xu, Song, Bi, Zhang, Zhang, Li, Wu, et~al.]{shao2024deepseekmath}
Zhihong Shao, Peiyi Wang, Qihao Zhu, Runxin Xu, Junxiao Song, Xiao Bi, Haowei Zhang, Mingchuan Zhang, YK~Li, Yang Wu, et~al.
\newblock Deepseekmath: Pushing the limits of mathematical reasoning in open language models.
\newblock \emph{arXiv preprint arXiv:2402.03300}, 2024.

\bibitem[Zhao et~al.(2026)Zhao, Kang, Feng, Levine, and Song]{zhao2026learning}
Xuandong Zhao, Zhewei Kang, Aosong Feng, Sergey Levine, and Dawn Song.
\newblock Learning to reason without external rewards.
\newblock In \emph{International Conference on Learning Representations}, volume 2026, pages 2548--2581, 2026.

\bibitem[Zuo et~al.(2026)Zuo, Zhang, Sheng, Qu, Cui, Zhu, Li, Long, Hua, Qi, et~al.]{zuo2026ttrl}
Yuxin Zuo, Kaiyan Zhang, Li~Sheng, Shang Qu, Ganqu Cui, Xuekai Zhu, Haozhan Li, Xinwei Long, Ermo Hua, Biqing Qi, et~al.
\newblock Ttrl: Test-time reinforcement learning.
\newblock \emph{Advances in Neural Information Processing Systems}, 38:\penalty0 131459--131483, 2026.

\bibitem[Li et~al.(2025{\natexlab{b}})Li, Zhao, So, Sun, and Li]{li2025model}
Mengqi Li, Lei Zhao, Anthony Man-Cho So, Ruoyu Sun, and Xiao Li.
\newblock A model can help itself: Reward-free self-training for llm reasoning.
\newblock \emph{arXiv preprint arXiv:2510.18814}, 2025{\natexlab{b}}.

\bibitem[Chen et~al.(2026)Chen, Chen, Li, and Huang]{pmlr-v306-chen26k}
Ruishuo Chen, Yu~Chen, Zhuoran Li, and Longbo Huang.
\newblock {P}ower{F}low: Unlocking the dual nature of {LLM}s via principled distribution matching.
\newblock In Tong Zhang, Miroslav Dudik, Martin Jaggi, Alekh Agarwal, Sharon Li, Dale Schuurmans, Jerry Zhu, Felix Berkenkamp, Hanze Dong, and Alberto Bietti, editors, \emph{Proceedings of the 43rd International Conference on Machine Learning}, volume 306 of \emph{Proceedings of Machine Learning Research}, pages 13700--13725. PMLR, 06--11 Jul 2026.
\newblock URL \url{https://proceedings.mlr.press/v306/chen26k.html}.

\bibitem[Gower and Legendre(1986)]{gower1986metric}
John~C Gower and Pierre Legendre.
\newblock Metric and euclidean properties of dissimilarity coefficients.
\newblock \emph{Journal of classification}, 3\penalty0 (1):\penalty0 5--48, 1986.

\bibitem[Chang et~al.(2025)Chang, Kim, Krumdick, Zadeh, Li, Tanner, and Iyyer]{chang2025bleuberi}
Yapei Chang, Yekyung Kim, Michael Krumdick, Amir Zadeh, Chuan Li, Chris Tanner, and Mohit Iyyer.
\newblock {BLEUBERI}: {BLEU} is a surprisingly effective reward for instruction following.
\newblock In \emph{The Thirty-ninth Annual Conference on Neural Information Processing Systems}, 2025.
\newblock URL \url{https://openreview.net/forum?id=bEYzeItoOH}.

\bibitem[Joshi et~al.(2017)Joshi, Choi, Weld, and Zettlemoyer]{triviaqa}
Mandar Joshi, Eunsol Choi, Daniel~S Weld, and Luke Zettlemoyer.
\newblock Triviaqa: A large scale distantly supervised challenge dataset for reading comprehension.
\newblock \emph{arXiv preprint arXiv:1705.03551}, 2017.

\bibitem[Yang et~al.(2018)Yang, Qi, Zhang, Bengio, Cohen, Salakhutdinov, and Manning]{yang2018hotpotqa}
Zhilin Yang, Peng Qi, Saizheng Zhang, Yoshua Bengio, William~W Cohen, Ruslan Salakhutdinov, and Christopher~D Manning.
\newblock Hotpotqa: A dataset for diverse, explainable multi-hop question answering.
\newblock \emph{arXiv preprint arXiv:1809.09600}, 2018.

\bibitem[Ho et~al.(2020)Ho, Nguyen, Sugawara, and Aizawa]{ho2020constructing}
Xanh Ho, Anh-Khoa~Duong Nguyen, Saku Sugawara, and Akiko Aizawa.
\newblock Constructing a multi-hop qa dataset for comprehensive evaluation of reasoning steps.
\newblock \emph{arXiv preprint arXiv:2011.01060}, 2020.

\bibitem[Trivedi et~al.(2022)Trivedi, Balasubramanian, Khot, and Sabharwal]{trivedi2022musique}
Harsh Trivedi, Niranjan Balasubramanian, Tushar Khot, and Ashish Sabharwal.
\newblock Musique: Multihop questions via single-hop question composition.
\newblock \emph{Transactions of the Association for Computational Linguistics}, 10:\penalty0 539--554, 2022.

\bibitem[Fan et~al.(2019)Fan, Jernite, Perez, Grangier, Weston, and Auli]{fan2019eli5}
Angela Fan, Yacine Jernite, Ethan Perez, David Grangier, Jason Weston, and Michael Auli.
\newblock Eli5: Long form question answering.
\newblock In \emph{Proceedings of ACL 2019}, 2019.

\bibitem[Kwiatkowski et~al.(2019)Kwiatkowski, Palomaki, Redfield, Collins, Parikh, Alberti, Epstein, Polosukhin, Devlin, Lee, et~al.]{kwiatkowski2019natural}
Tom Kwiatkowski, Jennimaria Palomaki, Olivia Redfield, Michael Collins, Ankur Parikh, Chris Alberti, Danielle Epstein, Illia Polosukhin, Jacob Devlin, Kenton Lee, et~al.
\newblock Natural questions: a benchmark for question answering research.
\newblock \emph{Transactions of the Association for Computational Linguistics}, 7:\penalty0 453--466, 2019.

\bibitem[Wang et~al.(2022)Wang, Yang, Huang, Jiao, Yang, Jiang, Majumder, and Wei]{wang2022text}
Liang Wang, Nan Yang, Xiaolong Huang, Binxing Jiao, Linjun Yang, Daxin Jiang, Rangan Majumder, and Furu Wei.
\newblock Text embeddings by weakly-supervised contrastive pre-training.
\newblock \emph{arXiv preprint arXiv:2212.03533}, 2022.

\bibitem[Petroni et~al.(2020)Petroni, Piktus, Fan, Lewis, Yazdani, Cao, Thorne, Jernite, Plachouras, Rockt"aschel, and Riedel]{fb_kilt}
Fabio Petroni, Aleksandra Piktus, Angela Fan, Patrick Lewis, Majid Yazdani, Nicola~De Cao, James Thorne, Yacine Jernite, Vassilis Plachouras, Tim Rockt"aschel, and Sebastian Riedel.
\newblock {KILT:} a {B}enchmark for {K}nowledge {I}ntensive {L}anguage {T}asks.
\newblock 2020.

\bibitem[Hu et~al.(2025)Hu, Zhang, Han, Jiang, Zhang, and Shum]{hu2025open}
Jingcheng Hu, Yinmin Zhang, Qi~Han, Daxin Jiang, Xiangyu Zhang, and Heung-Yeung Shum.
\newblock Open-reasoner-zero: An open source approach to scaling up reinforcement learning on the base model.
\newblock \emph{arXiv preprint arXiv:2503.24290}, 2025.

\bibitem[Yue et~al.(2024)Yue, Zhuang, Bai, Hui, Jagerman, Zeng, Qin, Wang, Wang, and Bendersky]{yue2024inference}
Zhenrui Yue, Honglei Zhuang, Aijun Bai, Kai Hui, Rolf Jagerman, Hansi Zeng, Zhen Qin, Dong Wang, Xuanhui Wang, and Michael Bendersky.
\newblock Inference scaling for long-context retrieval augmented generation.
\newblock \emph{arXiv preprint arXiv:2410.04343}, 2024.

\bibitem[Zhao et~al.(2024{\natexlab{b}})Zhao, Cao, Zhao, and Ou]{zhao2024empirical}
Yuetong Zhao, Hongyu Cao, Xianyu Zhao, and Zhijian Ou.
\newblock An empirical study of retrieval augmented generation with chain-of-thought.
\newblock In \emph{2024 IEEE 14th International Symposium on Chinese Spoken Language Processing (ISCSLP)}, pages 436--440. IEEE, 2024{\natexlab{b}}.

\bibitem[Chung et~al.(2024)Chung, Hou, Longpre, Zoph, Tay, Fedus, Li, Wang, Dehghani, Brahma, et~al.]{chung2024scaling}
Hyung~Won Chung, Le~Hou, Shayne Longpre, Barret Zoph, Yi~Tay, William Fedus, Yunxuan Li, Xuezhi Wang, Mostafa Dehghani, Siddhartha Brahma, et~al.
\newblock Scaling instruction-finetuned language models.
\newblock \emph{Journal of Machine Learning Research}, 25\penalty0 (70):\penalty0 1--53, 2024.

\bibitem[Burns et~al.(2023)Burns, Izmailov, Kirchner, Baker, Gao, Aschenbrenner, Chen, Ecoffet, Joglekar, Leike, et~al.]{burns2023weak}
Collin Burns, Pavel Izmailov, Jan~Hendrik Kirchner, Bowen Baker, Leo Gao, Leopold Aschenbrenner, Yining Chen, Adrien Ecoffet, Manas Joglekar, Jan Leike, et~al.
\newblock Weak-to-strong generalization: Eliciting strong capabilities with weak supervision.
\newblock \emph{arXiv preprint arXiv:2312.09390}, 2023.

\bibitem[Yao et~al.(2022)Yao, Zhao, Yu, Du, Shafran, Narasimhan, and Cao]{yao2022react}
Shunyu Yao, Jeffrey Zhao, Dian Yu, Nan Du, Izhak Shafran, Karthik Narasimhan, and Yuan Cao.
\newblock React: Synergizing reasoning and acting in language models.
\newblock \emph{arXiv preprint arXiv:2210.03629}, 2022.

\bibitem[Schick et~al.(2023)Schick, Dwivedi-Yu, Dess{\`\i}, Raileanu, Lomeli, Hambro, Zettlemoyer, Cancedda, and Scialom]{schick2023toolformer}
Timo Schick, Jane Dwivedi-Yu, Roberto Dess{\`\i}, Roberta Raileanu, Maria Lomeli, Eric Hambro, Luke Zettlemoyer, Nicola Cancedda, and Thomas Scialom.
\newblock Toolformer: Language models can teach themselves to use tools.
\newblock \emph{Advances in neural information processing systems}, 36:\penalty0 68539--68551, 2023.

\bibitem[Singh et~al.(2025)Singh, Ehtesham, Kumar, Khoei, and Vasilakos]{singh2025agentic}
Aditi Singh, Abul Ehtesham, Saket Kumar, Tala~Talaei Khoei, and Athanasios~V Vasilakos.
\newblock Agentic retrieval-augmented generation: A survey on agentic rag.
\newblock \emph{arXiv preprint arXiv:2501.09136}, 2025.

\bibitem[Papineni et~al.(2002)Papineni, Roukos, Ward, and Zhu]{papineni2002bleu}
Kishore Papineni, Salim Roukos, Todd Ward, and Wei-Jing Zhu.
\newblock Bleu: a method for automatic evaluation of machine translation.
\newblock In \emph{Proceedings of the 40th annual meeting of the Association for Computational Linguistics}, pages 311--318, 2002.

\bibitem[Chu et~al.(2025)Chu, Zhai, Yang, Tong, Xie, Schuurmans, Le, Levine, and Ma]{chu2025sft}
Tianzhe Chu, Yuexiang Zhai, Jihan Yang, Shengbang Tong, Saining Xie, Dale Schuurmans, Quoc~V Le, Sergey Levine, and Yi~Ma.
\newblock Sft memorizes, rl generalizes: A comparative study of foundation model post-training.
\newblock \emph{arXiv preprint arXiv:2501.17161}, 2025.

\bibitem[Courty et~al.(2024)Courty, Schmidt, Luccioni, Goyal-Kamal, MarionCoutarel, Feld, Lecourt, LiamConnell, Saboni, Inimaz, supatomic, Léval, Blanche, Cruveiller, ouminasara, Zhao, Joshi, Bogroff, de~Lavoreille, Laskaris, Abati, Blank, Wang, Catovic, Alencon, Michał Stęchły, Bauer, de~Araújo, JPW, and MinervaBooks]{benoit_courty_2024_11171501}
Benoit Courty, Victor Schmidt, Sasha Luccioni, Goyal-Kamal, MarionCoutarel, Boris Feld, Jérémy Lecourt, LiamConnell, Amine Saboni, Inimaz, supatomic, Mathilde Léval, Luis Blanche, Alexis Cruveiller, ouminasara, Franklin Zhao, Aditya Joshi, Alexis Bogroff, Hugues de~Lavoreille, Niko Laskaris, Edoardo Abati, Douglas Blank, Ziyao Wang, Armin Catovic, Marc Alencon, Michał Stęchły, Christian Bauer, Lucas Otávio~N. de~Araújo, JPW, and MinervaBooks.
\newblock mlco2/codecarbon: v2.4.1, May 2024.
\newblock URL \url{https://doi.org/10.5281/zenodo.11171501}.

\bibitem[Shorten and Khoshgoftaar(2019)]{shorten2019survey}
Connor Shorten and Taghi~M Khoshgoftaar.
\newblock A survey on image data augmentation for deep learning.
\newblock \emph{Journal of big data}, 6\penalty0 (1):\penalty0 60, 2019.

\bibitem[Ko et~al.(2015)Ko, Peddinti, Povey, and Khudanpur]{ko2015audio}
Tom Ko, Vijayaditya Peddinti, Daniel Povey, and Sanjeev Khudanpur.
\newblock Audio augmentation for speech recognition.
\newblock In \emph{Proc. Interspeech 2015}, pages 3586--3589, 2015.

\bibitem[Feng et~al.(2021)Feng, Gangal, Wei, Chandar, Vosoughi, Mitamura, and Hovy]{feng2021survey}
Steven~Y Feng, Varun Gangal, Jason Wei, Sarath Chandar, Soroush Vosoughi, Teruko Mitamura, and Eduard Hovy.
\newblock A survey of data augmentation approaches for nlp.
\newblock In \emph{Findings of the association for computational linguistics: ACL-IJCNLP 2021}, pages 968--988, 2021.

\bibitem[Wang et~al.(2023)Wang, Wei, Schuurmans, Le, Chi, Narang, Chowdhery, and Zhou]{wang2023selfconsistency}
Xuezhi Wang, Jason Wei, Dale Schuurmans, Quoc~V Le, Ed~H. Chi, Sharan Narang, Aakanksha Chowdhery, and Denny Zhou.
\newblock Self-consistency improves chain of thought reasoning in language models.
\newblock In \emph{The Eleventh International Conference on Learning Representations}, 2023.
\newblock URL \url{https://openreview.net/forum?id=1PL1NIMMrw}.

\bibitem[Guo et~al.(2025)Guo, Yang, Zhang, Song, Wang, Zhu, Xu, Zhang, Ma, Bi, et~al.]{guo2025deepseekr1}
Daya Guo, Dejian Yang, Haowei Zhang, Junxiao Song, Peiyi Wang, Qihao Zhu, Runxin Xu, Ruoyu Zhang, Shirong Ma, Xiao Bi, et~al.
\newblock Deepseek-r1: Incentivizing reasoning capability in llms via reinforcement learning.
\newblock \emph{arXiv preprint arXiv:2501.12948}, 2025.

\bibitem[Yu et~al.(2026)Yu, Zhang, Zhu, Yuan, Zuo, Yue, Dai, Fan, Liu, Liu, et~al.]{yu2026dapo}
Qiying Yu, Zheng Zhang, Ruofei Zhu, Yufeng Yuan, Xiaochen Zuo, Yu~Yue, Weinan Dai, Tiantian Fan, Gaohong Liu, Lingjun Liu, et~al.
\newblock Dapo: An open-source llm reinforcement learning system at scale.
\newblock \emph{Advances in Neural Information Processing Systems}, 38:\penalty0 113222--113244, 2026.

\bibitem[Liu et~al.(2025)Liu, Chen, Li, Qi, Pang, Du, Lee, and Lin]{liu2025understanding}
Zichen Liu, Changyu Chen, Wenjun Li, Penghui Qi, Tianyu Pang, Chao Du, Wee~Sun Lee, and Min Lin.
\newblock Understanding r1-zero-like training: A critical perspective.
\newblock \emph{arXiv preprint arXiv:2503.20783}, 2025.

\end{thebibliography}
\bibliographystyle{unsrtnat}


\newpage
\appendix
\section{Appendix}
\addtocontents{toc}{\protect\setcounter{tocdepth}{2}}
\makeatletter
\@starttoc{toc}
\makeatother

\subsection{Broader Impact}\label{apx:impact}
This work aims to improve the reliability of retrieval-augmented generation (RAG) systems by reducing inconsistency across semantically equivalent queries. More consistent RAG systems can increase user trust and reduce the risk of confusion or misinformation in high-stakes applications such as healthcare, finance, and legal assistance, where users expect the same guidance regardless of how a question is phrased. This becomes increasingly important as LLMs are deployed in agentic systems that chain reasoning with calls to external tools such as web search: an inconsistent answer at one step can propagate through subsequent actions, and retrieval is often provided by third-party services that cannot be modified, so robustness must come from the model itself.

At the same time, consistency is not correctness. A system that is consistently wrong may appear more authoritative than one that is inconsistent, and improved consistency may increase users' reliance on model outputs. Consistency optimization should therefore be paired with continued evaluation of accuracy, as we do throughout this work. In open-ended settings, enforcing agreement across paraphrases may also reduce legitimate diversity in responses, such as alternative valid perspectives, which may not always be desirable. We do not anticipate novel ethical risks beyond those commonly associated with large language models, but encourage future work on jointly addressing consistency, accuracy, and robustness in real-world deployments.

\subsection{Generating Paraphrased and Semantically Equivalent Queries}\label{apx:paraphrase}
For each query $q_0$, we use LLaMA-3.1-70B  to generate $n$ paraphrases $\mathcal{P}(q_0) = \{p_1, \dots, p_n\}$. To ensure answerability, we provide the ground truth answer as part of the prompt and instruct the model to generate paraphrases that preserve the exact meaning such that each paraphrase can be answered in the same way. In our early experiments, we found paraphrases generated without the gold answer degraded the quality of paraphrases generated: they remained similar to the original query but slightly changed the implied answer or introduced new answerable variants. This directly harmed the validity of our consistency evaluation. Including the gold answer anchored the paraphrase generation to the same underlying fact. For example, given the original question ``\textit{What nationality was Person X?}'' (answer: \textit{American}), a paraphrase generated without the gold answer was ``\textit{What was the ethnic background of Person X?}'' Providing the gold answer resolved such issues, and in practice we observed no cases where the gold answer leaked into the paraphrases.

\begin{table}[h]
\centering
\caption{\small
\textbf{Effect of gold answers on paraphrase quality} (HotpotQA).
Accuracy on original queries and paraphrased queries generated with and without access to the gold answer.
Including the gold answer during paraphrase generation preserves semantic equivalence, while omitting it
leads to substantial accuracy degradation.
}
\small
\begin{tabular}{l|l|ccc}
\toprule
\textbf{Setting} & \textbf{Query Type} & \textbf{EM} & \textbf{F1} & \textbf{RM} \\
\midrule
\multirow{1}{*}{--} 
& Original Queries        & 37.0 & 44.1 & 42.0 \\
\midrule
With Gold Answer 
& Paraphrased Queries     & 36.4 & 43.5 & 42.4 \\
\midrule
Without Gold Answer 
& Paraphrased Queries     & 27.4 & 36.1 & 35.4 \\
\bottomrule
\end{tabular}
\label{tab:gold_answer_para}
\end{table}

To further demonstrate this, we generate paraphrases on HotpotQA both with and without ground truth and compare accuracy across original queries and paraphrased queries. As shown in Table~\ref{tab:gold_answer_para}, with the gold answer included, paraphrased queries maintain nearly identical accuracy to the original queries (e.g., EM drops from 37.0 → 36.4). In contrast, when the gold answer is omitted, accuracy on paraphrased queries drops sharply (EM: 37.0 → 27.4), indicating that many paraphrases drift semantically. Importantly, this does not affect the model being evaluated. Paraphrases are generated offline, and the ground truth is never shown to the RAG system.

We note that for long-form QA tasks (e.g., ELI5, NQ) we do not include ground truth during paraphrase generation, and this issue does not arise. See prompt used for  short-form  and long form QA tasks below:

\begin{tcolorbox}[title=\texttt{Paraphrasing -- Short-form QA}, colback=gray!5, colframe=black!50, fonttitle=\bfseries]
\small
You are given an input sentence. Your task is to generate \texttt{{n}} diverse paraphrases of this sentence.
You can paraphrase by using synonyms, changing sentence structure, or rephrasing in any other way,
but each paraphrase should preserve the original meaning.
Each paraphrase you create must be answerable by the exact same answer provided below.

Format your output as follows:

\texttt{<paraphrase1>} paraphrased sentence 1 \texttt{</paraphrase1>}

\texttt{<paraphrase2>} paraphrased sentence 2 \texttt{</paraphrase2>}

...

\texttt{<paraphrase{n}>} paraphrased sentence \texttt{{n}} \texttt{</paraphrase{n}>}

\textbf{Input sentence:} \texttt{\{sentence\}}
\textbf{Required answer:} \texttt{\{answer\}}

Please return only the paraphrases in the specified format.
\end{tcolorbox}

\begin{tcolorbox}[title=\texttt{Paraphrasing -- Long-form QA}, colback=gray!5, colframe=black!50, fonttitle=\bfseries]
\small
You are given an input question sentence. Your task is to generate \texttt{{n}} diverse paraphrases of this question.
You can paraphrase by using synonyms, changing sentence structure, or rephrasing in any other way,
but each paraphrase should preserve the original question meaning and lead to similar answers.

Format your output as follows:

\texttt{<paraphrase1>} paraphrased sentence 1 \texttt{</paraphrase1>}

\texttt{<paraphrase2>} paraphrased sentence 2 \texttt{</paraphrase2>}

...

\texttt{<paraphrase{n}>} paraphrased sentence \texttt{{n}} \texttt{</paraphrase{n}>}

\textbf{Input question:} \texttt{\{sentence\}}

Please return only the paraphrases in the specified format.
\end{tcolorbox}

\subsection{Measuring Information  Consistency Using LLM-as-a-Judge}\label{app:llm-judge}

To measure information semantic consistency between generated outputs across paraphrased queries, we employ LLM-as-a-judge  as part of our evaluation pipeline. This judge is designed to determine whether two answers convey the same core information, even if they differ in surface form. For all LLaMA-3.1-8B experiments, we use LLaMA-3.3-70B as the evaluator. For all Qwen-2.5-3B experiments, we use GPT-4o as the evaluator. An example evaluation prompt is shown below:

\begin{tcolorbox}[title=\texttt{Consistency Evaluation -- Long-form QA}, colback=gray!5, colframe=black!50, fonttitle=\bfseries]
\small
You are an evaluator assessing two different answers that were generated for the same question. Your task is to determine if the two answers are consistent with each other.

Consider them consistent if they present the same core information about the underlying question.  
Consider them inconsistent if they provide different information, present different facts, or address the underlying question in different ways.

Based on this, reply with only \texttt{yes} or \texttt{no}. Do not output anything else.

\textbf{Answer 1:} \texttt{\{output\_i\}}

\textbf{Answer 2:} \texttt{\{output\_j\}}

\textbf{Are these two answers consistent? (yes/no). Response:}
\end{tcolorbox}

\vspace{0.75em}

\begin{tcolorbox}[title=\texttt{Consistency Evaluation -- Short-form QA}, colback=gray!5, colframe=black!50, fonttitle=\bfseries]
\small
Given the following two outputs sentences, reply with \texttt{yes} if the two sentences are consistent with each other, or \texttt{no} if they are not. Do not output anything else.

\textbf{Sentence 1:} \texttt{\{output\_i\}}

\textbf{Sentence 2:} \texttt{\{output\_j\}}

\textbf{Are these sentences consistent? (yes/no). Response:}
\end{tcolorbox}

\begin{table}[t]
\centering
\caption{\small
\textbf{Accuracy across datasets and query variants} (LLaMA-3.1-8B). We report accuracy for original queries, synthetically generated paraphrased queries, and paraphrased queries with fixed retrieval. Across all settings, accuracy remains relatively similar, indicating that paraphrasing and retrieval shifts have limited effect on final answer correctness on average. See result for Qwen-2.5-3B model in Table~\ref{tab:accuracy-table-qwen}.
}

\small
\begin{tabular}{l|ccc|ccc|ccc}
\toprule
\multicolumn{10}{c}{\textbf{Short-form \& Multi-hop QA: Accuracy (\%)}} \\
\midrule
\multirow{2}{*}{\textbf{Dataset}} 
& \multicolumn{3}{c}{\textbf{Original Queries}} 
& \multicolumn{3}{c}{\textbf{Paraphrased Queries}} 
& \multicolumn{3}{c}{\textbf{Paraphrased (Fixed Docs)}} \\
\cmidrule(lr){2-4}\cmidrule(lr){5-7}\cmidrule(lr){8-10}
& EM & F1 & RM & EM & F1 & RM & EM & F1 & RM \\
\midrule
TriviaQA   & 56.0 & 66.1 & 74.0 & 55.0 & 64.4 & 73.3 & 58.7 & 67.3 & 75.0 \\
HotpotQA  & 37.0 & 44.1 & 42.0 & 36.4 & 43.5 & 42.4 & 33.7 & 40.7 & 39.4 \\
2Wiki      & 28.0 & 33.9 & 37.0 & 25.9 & 31.3 & 32.7 & 26.9 & 31.7 & 33.3 \\
MuSiQue    &  8.0 & 15.3 & 12.0 &  8.3 & 14.1 & 11.0 & 11.0 & 17.5 & 15.0 \\
\bottomrule
\end{tabular}

\vspace{6pt}

\begin{tabular}{l|cc|cc|cc}
\toprule
\multicolumn{7}{c}{\textbf{Long-form QA: Accuracy (\%)}} \\
\midrule
\multirow{2}{*}{\textbf{Dataset}} 
& \multicolumn{2}{c}{\textbf{Original Queries}} 
& \multicolumn{2}{c}{\textbf{Paraphrased Queries}} 
& \multicolumn{2}{c}{\textbf{Paraphrased (Fixed Docs)}} \\
\cmidrule(lr){2-3}\cmidrule(lr){4-5}\cmidrule(lr){6-7}
& ROUGE & LLM-Acc & ROUGE & LLM-Acc & ROUGE & LLM-Acc \\
\midrule
ELI5 & 21.9 & 74.0 & 20.7 & 71.3 & 20.8 & 70.3 \\
NQ  &26.9 & 61.0  &25.2 & 58.8 &25.2 & 57.9  \\
\bottomrule
\end{tabular}
\label{tab:accuracy-table-llama}
\end{table}

\begin{table}[t]
\centering
\caption{\small 
\textbf{Disentangling sources of inconsistency in RAG systems} (Qwen-2.5-3B). Retriever consistency is low across datasets, suggesting that paraphrased queries often retrieve non-overlapping documents. This introduces context variability that is reflected in the end-to-end consistency scores. Fixing retrieval improves consistency, but variation remains, revealing the generator’s sensitivity to input phrasing even with identical evidence.}
\small
\begin{tabular}{l|cc|cc|c}
\toprule
\multirow{2}{*}{\textbf{Dataset}} 
& \multicolumn{2}{c}{\textbf{End-to-End Consistency}} 
& \multicolumn{2}{c}{\textbf{Generator (LLM) Consistency}} 
& \textbf{Retriever Consistency} \\
\cmidrule(lr){2-3}\cmidrule(lr){4-5}\cmidrule(lr){6-6}
& Lexical & Inform. & Lexical & Inform. & Jaccard Overlap \\
\midrule
TriviaQA   & 47.9 & 73.0 & 58.6 & 87.5 & 32.5 \\
HotpotQA  & 32.7 & 63.6 & 48.0 & 77.3 & 46.0 \\
2Wiki      & 32.3 & 62.6 & 44.6 & 70.7 & 52.4 \\
MuSiQue    & 25.7 & 49.5 & 45.7 & 67.3 & 36.6 \\
ElI5      & 6.6  & 35.3 & 14.4 & 62.3 & 27.1 \\
\bottomrule
\end{tabular}
\label{tbl:consistency-table_qwen}
\end{table}

\begin{table}[t]
\centering
\caption{\small
\textbf{Accuracy across datasets and query variants} ({Qwen-2.5-3B}). We report accuracy for original queries, synthetically generated paraphrased queries, and paraphrased queries with fixed retrieval. Across all settings, accuracy remains relatively similar, indicating that paraphrasing and retrieval shifts have limited effect on final answer correctness on average.
}

\small
\begin{tabular}{l|ccc|ccc|ccc}
\toprule
\multicolumn{10}{c}{\textbf{Short-form \& Multi-hop QA: Accuracy (\%)}} \\
\midrule
\multirow{2}{*}{\textbf{Dataset}} 
& \multicolumn{3}{c}{\textbf{Original Queries}} 
& \multicolumn{3}{c}{\textbf{Paraphrased Queries}} 
& \multicolumn{3}{c}{\textbf{Paraphrased (Fixed Docs)}} \\
\cmidrule(lr){2-4}\cmidrule(lr){5-7}\cmidrule(lr){8-10}
& EM & F1 & RM & EM & F1 & RM & EM & F1 & RM \\
\midrule
TriviaQA   & 42.0 & 50.7 & 58.0 & 46.3 & 54.1 & 64.3 & 43.0 & 51.1 & 62.3 \\
HotpotQA  & 20.0 & 28.3 & 37.0 & 20.9 & 27.7 & 38.4 & 18.2 & 26.4 & 38.4 \\
2Wiki      & 13.0 & 20.4 & 36.0 & 11.1 & 19.8 & 34.7 & 12.5 & 20.2 & 32.3 \\
MuSiQue    &  4.0 & 10.0 & 9.0  &  6.0 & 9.9  & 8.0  & 5.3  & 9.7  & 7.7 \\
\bottomrule
\end{tabular}

\vspace{6pt}

\begin{tabular}{l|cc|cc|cc}
\toprule
\multicolumn{7}{c}{\textbf{Long-form QA: Accuracy (\%)}} \\
\midrule
\multirow{2}{*}{\textbf{Dataset}} 
& \multicolumn{2}{c}{\textbf{Original Queries}} 
& \multicolumn{2}{c}{\textbf{Paraphrased Queries}} 
& \multicolumn{2}{c}{\textbf{Paraphrased (Fixed Docs)}} \\
\cmidrule(lr){2-3}\cmidrule(lr){4-5}\cmidrule(lr){6-7}
& ROUGE & LLM-Acc & ROUGE & LLM-Acc & ROUGE & LLM-Acc \\
\midrule
ElI5& 22.1 & 38.0 & 21.5 & 37.3 & 20.8 & 35.7 \\
\bottomrule
\end{tabular}
\label{tab:accuracy-table-qwen}
\end{table}

\begin{table}[t]
\centering
\caption{\small \textbf{Comparison between Con-RAG and Baselines} (Short-form QA, Qwen-2.5-3B). Lexical consistency measured via BLEU score while information consistency measured using an LLM-as-a-judge. Con-RAG is trained with a group-similarity reward plus an accuracy reward (no KL), and consistently yields higher end-to-end and generator-only consistency while also improving accuracy over original queries.}
\small
\begin{tabular}{ll|ccc|cc|cc}
\toprule
 & & \multicolumn{3}{c}{\textbf{Accuracy (\%)}} 
 & \multicolumn{2}{c}{\shortstack[c]{\textbf{End-to-End}\\\textbf{Consistency (\%)}}} 
 & \multicolumn{2}{c}{\shortstack[c]{\textbf{Generator (LLM)}\\\textbf{Consistency (\%)}}} \\
\cmidrule(lr){3-5} \cmidrule(lr){6-7} \cmidrule(lr){8-9}
\textbf{Dataset} & \textbf{Method} 
& EM & F1 & RM 
& Lexical & Inform. 
& Lexical & Inform. \\
\midrule
\multirow{5}{*}{TriviaQA}
& RAG          & 42.0 & 50.7 & 58.0 & 47.9 & 73.0 & 58.6 & 87.5 \\
& DRAG         & 42.0 & 50.7 & 58.0 & 47.9 & 73.5 & 58.6 & 84.7 \\
& CoT-RAG      & 37.0 & 44.5 & 61.0 & 41.1 & 72.3 & 52.2 & 82.3 \\
& SFT          & 35.0 & 40.4 & 43.0 & 53.3 & 72.2 & 73.4 & 85.0 \\
& \textbf{Con-RAG} & \textbf{60.0} & \textbf{66.0} & \textbf{68.0} & \textbf{67.1} & \textbf{81.8} &\textbf{ 80.5} & \textbf{89.5} \\
\midrule
\multirow{5}{*}{HotpotQA}
& RAG          & 20.0 & 28.3 & 37.0 & 32.7 & 63.6 & 48.0 & 77.3 \\
& DRAG         & 20.0 & 28.3 & 37.0 & 32.7 & 64.3 & 48.0 & 76.8 \\
& CoT-RAG      & 29.0 & 32.8 & 37.0 & 28.5 & 63.6 & 35.7 & 71.2 \\
& SFT          & 30.0 & 35.4 & 32.0 & 63.3 & 77.1 & 74.5 & 85.7 \\
& \textbf{Con-RAG} & \textbf{36.0} & \textbf{43.1} & \textbf{38.0} & \textbf{64.6} & \textbf{78.2 }& \textbf{77.8} & \textbf{86.7} \\
\midrule
\multirow{5}{*}{MuSiQue}
& RAG          &  4.0 & 10.0 &  9.0 & 25.7 & 49.5 & 45.7 & 67.3 \\
& DRAG         &  4.0 & 10.0 &  9.0 & 25.7 & 50.3 & 45.7 & 69.2 \\
& CoT-RAG      &  5.0 & 10.9 &  9.0 & 18.1 & 52.0 & 26.5 & 62.0 \\
& SFT          & 25.0 & 30.6 & 27.0 & 57.7 & 65.3 & 69.8 & 77.2 \\
& \textbf{Con-RAG} &  \textbf{27.0} & \textbf{31.9} &  \textbf{28.1} & \textbf{69.8} & \textbf{70.1} & \textbf{70.4} & \textbf{82.0 }\\
\midrule
\multirow{5}{*}{2Wiki}
& RAG          & 13.0 & 20.4 & 36.0 & 32.3 & 62.6 & 44.6 & 70.7 \\
& DRAG         & 13.0 & 20.4 & 36.0 & 32.3 & 63.0 & 44.6 & 70.9 \\
& CoT-RAG      & 23.0 & 27.0 & 30.0 & 23.5 & 62.3 & 32.7 & 67.0 \\
& SFT          & \textbf{37.0} & \textbf{38.9} & \textbf{38.0} & \textbf{70.9} & 75.8 & \textbf{84.8} & 86.9 \\
& \textbf{Con-RAG} & \textbf{37.0} & 38.4 & 37.0 & 68.2 & \textbf{76.6} & \textbf{84.8} & \textbf{89.1} \\
\bottomrule
\end{tabular}
\label{tab:main-results-qwen}
\end{table}

\begin{table}[t]
\centering
\caption{\small \textbf{Comparison between Con-RAG and baselines} (Long-form QA). Con-RAG is trained using only the group-similarity reward with a small KL regularizer (no accuracy supervision). Despite no ground-truth, it achieves the best end-to-end and generator consistency and also maintains answer quality over baselines, whereas SFT on reference answers underperforms in this open-ended setting (Qwen-2.5-3B).}
\small
\begin{tabular}{ll|cc|cc|cc}
\toprule
 & & \multicolumn{2}{c}{\textbf{Accuracy (\%)}} 
 & \multicolumn{2}{c}{\shortstack[c]{\textbf{End-to-End}\\\textbf{Consistency (\%)}}} 
 & \multicolumn{2}{c}{\shortstack[c]{\textbf{Generator (LLM)}\\\textbf{Consistency (\%)}}} \\
\cmidrule(lr){3-4} \cmidrule(lr){5-6} \cmidrule(lr){7-8}
\textbf{Dataset} & \textbf{Method} 
& ROUGE & LLM-Acc 
& Lexical & Inform. 
& Lexical & Inform. \\
\midrule
\multirow{7}{*}{{ELI5}}
& RAG         & 22.1 & 38.0 & 6.6  & 35.3 & 14.4 & 62.3 \\
& DRAG        & 22.1 & 38.0 & 6.6  & 35.3 & 14.4 & 63.8 \\
& CoT\text{-}RAG & 21.1 & 36.0 & 4.9  & 34.0 & 9.6  & 55.5 \\
& SFT         & \textbf{24.3} & 36.0 & 5.4  & 17.2 & 7.0  & 19.0 \\
& CoG   &23.6      & 40.2 & 6.2 & 20.4 & 10.3 &27.3    \\
& CAT   &23.8      & 43.6 & 5.8 & 25.5&  10.1 & 29.3   \\
& \textbf{Con\text{-}RAG} & 22.6 & \textbf{58.0} & \textbf{9.3} & \textbf{42.8} & \textbf{17.9} & \textbf{67.5} \\
\bottomrule
\end{tabular}
\end{table}

\subsection{BLEU: An $n$-gram Based Evaluation Metric}\label{apx:bleu}

The BLEU (Bilingual Evaluation Understudy~\cite{papineni2002bleu}) score is a standard metric for assessing the quality of machine translation. It quantifies the degree of overlap between a system-generated translation and one or more human reference translations. The score relies on modified $n$-gram precision ($n \in \{1,2,3,4\}$), together with a brevity penalty (BP) that discourages excessively short outputs:
\[
\text{BLEU} = \text{BP} \cdot \exp \left( \sum_{n=1}^{N} w_n \log p_n \right), 
\quad 
\text{BP} = 
\begin{cases} 
1 & \text{if } c > r, \\
\exp\!\left(1 - \tfrac{r}{c}\right) & \text{if } c \leq r,
\end{cases}
\]
where $p_n$ denotes the modified $n$-gram precision, $w_n$ are the associated weights, $c$ is the length of the candidate translation, and $r$ is the length of the closest reference. When multiple references are given, BLEU counts $n$-gram matches against all references and uses the maximum match count for each $n$-gram. Each $n$-gram level in BLEU captures progressively deeper aspects of linguistic quality. Unigrams ($n{=}1$) assess word choice or adequacy, indicating whether the candidate includes the correct content words. Bigrams ($n{=}2$) begin to reflect local fluency by capturing short-range word ordering. Trigrams ($n{=}3$) provide signals about phrase-level coherence, identifying whether multi-word chunks align with natural phrasing. 4-grams ($n{=}4$) enforce sentence-level fluency by requiring longer, contiguous sequences to match the reference.

\subsection{Training Configuration}\label{apx:training-config}
All Con-RAG models are trained with full-parameter fine-tuning in TRL on 8$\times$NVIDIA A100 GPUs, using DeepSpeed ZeRO-3 and bf16 mixed precision. We use AdamW with a learning rate of $(10^{-5},10^{-6})$ and a linear schedule without warmup, train for 4 epochs, and set the GRPO clipping parameter to $\epsilon{=}0.2$.  Rewards use $(\kappa=3, s=1)$. Short-form and multi-hop QA use BLEU-1 plus a token-F1 accuracy reward ($\alpha{=}\gamma{=}1$, $\beta{=}0$); long-form QA uses BLEU-2 with consistency only ($\gamma{=}0$, $\beta{=}0.05$). Training rollouts are sampled at temperature 1.0 and top-$p$ 1.0, with a maximum completion length of 16 tokens for short-form and 64 tokens for long-form QA; evaluation uses greedy decoding with $n{=}5$ paraphrases per query. Full configurations and scripts are available at \url{https://github.com/CapitalOne-Research/con-rag}.

\subsection{Propagation of Retrieval Inconsistency to End-to-End Consistency}\label{apx:jaccard-bins}

\textit{Retrieval variation propagates into output inconsistency, and Con-RAG's gains concentrate where it is most prevalent.} The fixed-retrieval experiment in Table~\ref{tbl:consistency-table} shows that generator sensitivity remains when retrieval is controlled, but does not by itself explain how much retriever inconsistency propagates into final output inconsistency. To quantify this relationship, we conduct a per-query analysis on NQ. For each canonical query, we compute (i) retriever Jaccard overlap across its paraphrase set, (ii) generator consistency under fixed retrieval, and (iii) end-to-end consistency, then bin queries by retrieval overlap and report conditional consistency (see Table~\ref{tab:jaccard-bins}). Let $\Delta$ be the difference between generator consistency, where every paraphrase is conditioned on identical evidence, and end-to-end consistency, where each paraphrase retrieves its own documents. $\Delta$ measures the share of output inconsistency that disappears once evidence is equalized.

For standard RAG, end-to-end information consistency rises from 41.5\% to $64.9\%$ as retrieval overlap increases, while fixed-retrieval generator consistency remains nearly unchanged ($67.0 - 69.2\%$). Accordingly, the retrieval-associated gap decreases from $25.5$ points under low overlap to only 4.3 points under high overlap. This directly shows that retrieval variation propagates into output inconsistency when paraphrases retrieve different evidence.

Con-RAG provides its largest improvement in the low-overlap setting, increasing end-to-end consistency from 41.5\% to 55.2\% and reducing the gap from 25.5 to 19.6 points. Low and medium-overlap queries constitute 76\% of the evaluation set, indicating that Con-RAG is most beneficial precisely where retrieval variation is prevalent. In the high-overlap bin, where retrieval variation is minimal, RAG and Con-RAG perform comparably on end-to-end consistency (64.9 vs. 67.5), as expected, there is little retrieval-induced variability left to absorb, and Con-RAG's advantage there appears at the generator level (69.2 → 76.1). Con-RAG also improves fixed-retrieval generator consistency in every bin.

\begin{table}[t]
\centering
\caption{\small
\textbf{Retrieval-induced inconsistency propagation} (NQ, LLaMA-3.1-8B).
Queries are binned by retriever consistency (Jaccard overlap across the paraphrase set). We report  LLM-Judge consistency, and the propagation gap $\Delta = \text{Gen.} - \text{E2E}$, i.e., the share of output inconsistency that disappears once evidence is equalized across paraphrases. For RAG, $\Delta$ shrinks from $25.5$ to $4.3$ points as retrieval overlap increases, showing that retrieval variation propagates into output inconsistency. Con-RAG's largest end-to-end gains occur in the low-overlap regime, where retrieval variation is the dominant source of inconsistency.}
\small
\begin{tabular}{l|c|l|cc|c}
\toprule
\textbf{Retriever Jaccard} & \textbf{\% Queries} & \textbf{Method} & \textbf{E2E Cons.} & \textbf{Gen. Cons.} & \textbf{$\Delta$ (Gen.$-$E2E)} \\
\midrule
\multirow{2}{*}{0.0--0.2 (low)}  & \multirow{2}{*}{31\%} & RAG     & 41.5 & 67.0 & 25.5 \\
                                 &                       & Con-RAG & \textbf{55.2} & \textbf{74.8} & \textbf{19.6} \\
\midrule
\multirow{2}{*}{0.2--0.5 (mid)}  & \multirow{2}{*}{45\%} & RAG     & 54.0 & 68.4 & 14.4 \\
                                 &                       & Con-RAG & \textbf{59.8} & \textbf{75.6} & 15.8 \\
\midrule
\multirow{2}{*}{0.5--1.0 (high)} & \multirow{2}{*}{24\%} & RAG     & 64.9 & 69.2 & 4.3 \\
                                 &                       & Con-RAG & \textbf{67.5} & \textbf{76.1} & 8.6 \\
\bottomrule
\end{tabular}
\label{tab:jaccard-bins}
\end{table}

\subsection{Comparison Between PS-GRPO and Standard GRPO}\label{apx:com_PSGRPO_GRPO}
PS-GRPO does not require any ground-truth references to compute rewards. This is a central property of PS-GRPO: the model improves information consistency purely from group, without using reference ground truth answers. This makes the method applicable to long-form and open-ended QA settings where single gold answers are unavailable or insufficient. We compare with standard GRPO, with  BLEU between the generated output and ground-truth answers as reward. Empirically, PS-GRPO outperforms GRPO with ground-truth on both consistency and accuracy (see results in  Table~\ref{tab:psgrpo-vs-grpo}). The learning signal from mutual agreement across paraphrases seem to be more effective than simply pushing each response toward a single reference.

\begin{table}[h]
\centering
\caption{\small
\textbf{Comparison between PS-GRPO and standard GRPO on Natural Questions (NQ).}
Standard GRPO uses BLEU reward between generated output and ground-truth answers, while PS-GRPO (Con-RAG)
uses only group similarity rewards without ground-truth supervision. PS-GRPO outperforms GRPO with ground-truth on both consistency and accuracy.}
\small
\begin{tabular}{l|cc|cc|cc}
\toprule
\multirow{2}{*}{\textbf{Method}} 
& \multicolumn{2}{c}{\textbf{Accuracy (\%)}} 
& \multicolumn{2}{c}{\textbf{End-to-End Cons. (\%)}} 
& \multicolumn{2}{c}{\textbf{Generator Cons. (\%)}} \\
\cmidrule(lr){2-3}\cmidrule(lr){4-5}\cmidrule(lr){6-7}
& ROUGE & LLM-Acc 
& Lexical & Inform. 
& Lexical & Inform. \\
\midrule
GRPO (BLEU reward -- \\ ground truth supervision ) 
& 37.4 & 52.0 & 19.2 & 47.5 & 29.6 & 60.5 \\
PS-GRPO (Con-RAG -- \\ no ground truth supervision) 
& 23.5 & \textbf{59.0} & \textbf{22.8} & \textbf{59.3} & \textbf{32.3} & \textbf{75.5} \\
\bottomrule
\end{tabular}
\label{tab:psgrpo-vs-grpo}
\end{table}

\subsection{Con-RAG outperforming SFT for Short-Form QA}\label{apx:sft}
SFT treats each paraphrase-answer pair independently and minimizes token-level cross-entropy with a single reference answer (or ground truth). This encourages the model to reproduce the surface form of that reference and penalizes any deviation (even when multiple phrasings are equally correct). For instance, a date like “November 16th, 2025” may be expressed accurately in several valid forms (“\textit{Nov. 16, 2025},” “\textit{16 November 2025,}” etc.), yet SFT penalizes all but the exact reference. This also explains why the F1-based accuracy reward was most effective in our experiments compared to Exact match (EM) and Relaxed Match (RM) as it provides a more flexible alignment signal that tolerates variations in phrasing (see Table~\ref{tab:abl-acc}). In contrast, Con-RAG explicitly optimizes a group-level objective over paraphrase sets. Outputs are rewarded not only for matching the ground truth but also for being consistent with outputs generated for other paraphrases of the same underlying question. This means that alternative answer formulations can still receive positive reward if they remain consistent across paraphrases, rather than being penalized for surface-level differences.

One could have a variant of our approach for SFT which has a loss term that penalizes inconsistency across paraphrases. However, recent evidence further supports our design choice of using an RL approach rather than SFT to optimize consistency: SFT Memorizes, RL Generalizes~\cite{chu2025sft} shows that SFT largely encourages memorization of surface patterns, causing models to overfit to the exact phrasing and structure of training examples, while RL, with outcome-level rewards, learns transferable rules that robustly generalize to unseen variants. Thus, Con-RAG’s gains over SFT align with the broader finding that RL is inherently better suited for learning structures that must remain stable across input variations.

\subsection{Training Cost–Performance Trade-offs under $(\kappa, s)$ Subsampling} \label{apx:compute_ablation}To assess the computational trade-offs introduced by the relaxed group-similarity reward, we performed a controlled sweep over the subsampling parameters $\kappa$ (number of paraphrases sampled) and $s$ (rollouts per paraphrase) on the NQ (Natural Questions) dataset while logging CPU/GPU energy and power using the CodeCarbon python package~\cite{benoit_courty_2024_11171501}. Results are summarized in Table~\ref{tab:nq-kappa-s}. We find that Con-RAG’s performance is stable across a wide range of $(\kappa, s)$ settings: increasing $\kappa$ and $s$ yields only mild fluctuations in accuracy and consistency metrics. However, the computational footprint grows noticeably with larger subsampling budgets. These results highlight that the reward signal is not highly sensitive to these hyperparameters, meaning practitioners can safely adopt low-cost settings which already provide strong consistency gains at a fraction of the compute. For high-resource settings, larger ($\kappa$, $s$) provide marginal additional improvements but at a noticeably higher cost, defining a compute-quality frontier for deployment scenarios with fixed budgets.

\begin{table}[t]
\centering
\setlength{\tabcolsep}{4pt}
\caption{\small
\textbf{Natural Questions (NQ) results across $(\kappa, s)$ settings.}
We report accuracy, consistency, and computational cost while varying
the number of sampled paraphrases $\kappa$ and rollouts per paraphrase $s$.
}
\small
\begin{tabular}{l|cc|cc|cc|cc|cc}
\toprule
\multirow{2}{*}{\textbf{Dataset}} 
& \multicolumn{2}{c}{\textbf{Params}} 
& \multicolumn{2}{c}{\textbf{Accuracy}} 
& \multicolumn{2}{c}{\textbf{E2E Consistency}} 
& \multicolumn{2}{c}{\textbf{Gen. Consistency}} 
& \multicolumn{2}{c}{\textbf{Compute}} \\
\cmidrule(lr){2-3}
\cmidrule(lr){4-5}
\cmidrule(lr){6-7}
\cmidrule(lr){8-9}
\cmidrule(lr){10-11}
& $\kappa$ & $s$ 
& ROUGE & LLM-Acc 
& Lex. & Inform. 
& Lex. & Inform. 
& CPU (W) & GPU (W) \\
\midrule
NQ 
& 1 & 1 & 25.2 & 60.0 & 21.3 & 60.5 & 30.7 & 70.8 & 118 & 441 \\
& 2 & 1 & 24.1 & 59.0 & 31.3 & 59.3 & 42.7 & 74.0 & 120 & 457 \\
& 5 & 1 & 24.3 & 60.0 & 28.4 & 59.3 & 38.5 & 73.0 & 123 & 462 \\
& 5 & 5 & 25.6 & 63.0 & 23.5 & 59.8 & 33.9 & 76.2 & 135 & 492 \\
\bottomrule
\end{tabular}
\label{tab:nq-kappa-s}
\end{table}
\subsection{Semantic Similarity Reward: BLEU vs.\ BERTScore}\label{apx:bertscore}
Our default group similarity reward is instantiated with BLEU, a lexical metric. A natural question is whether a semantic similarity function yields better information consistency, and whether the gains of Con-RAG are driven by surface-level lexical convergence (i.e., the policy concentrating onto canonical, high-overlap responses) rather than agreement in information content. To study this, we replace BLEU with BERTScore as the similarity function $\mathrm{sim}(\cdot,\cdot)$ in the group similarity reward on NQ (LLaMA-3.1-8B), keeping all other training settings identical to our default long-form configuration ($\kappa{=}3$, $s{=}1$). As in Appendix~\ref{apx:compute_ablation}, we log CPU and GPU power using the CodeCarbon python package~\cite{benoit_courty_2024_11171501}. Results are summarized in Table~\ref{tab:bertscore-reward}.

\textit{Information consistency is not a by-product of lexical convergence.} BERTScore achieves comparable information consistency but with lower lexical similarity scores compared to BLEU, which is expected, since BERTScore permits semantically equivalent but lexically diverse outputs. With BERTScore, lexical agreement drops markedly (E2E $22.8 \rightarrow 18.5$; Gen. $32.3 \rightarrow 27.4$), while LLM-Judge consistency and accuracy are essentially unchanged (E2E $59.3 \rightarrow 60.1$; Gen. $75.5 \rightarrow 74.8$; LLM-Acc $59.0 \rightarrow 60.0$). If the gains of Con-RAG were driven by surface-level mode concentration, they should depend on a reward that rewards surface overlap. Instead, when the reward tolerates lexical diversity, the model retains phrasing diversity while preserving content agreement. This is consistent with our observation that Con-RAG's LLM-Judge consistency substantially exceeds its lexical consistency across datasets (Tables~\ref{tab:main-results} \& \ref{tab:main-results-longform}).

\textit{Compute trade-off.} The comparable performance comes at a significantly higher computational cost: GPU power increases from 459W to 629W (37\%) and CPU power from 121W to 141W, since each comparison requires a BERT encoder forward pass. In contrast, BLEU is purely string-based and each comparison is negligible in cost. This matters in our setting, where group similarity rewards must be computed hundreds of times per optimization step. We view this as a cost-effectiveness spectrum: BLEU offers the best efficiency-to-performance ratio; BERTScore trades compute for tolerance of lexical diversity; and an LLM-as-a-judge would enable richer, prompt-specified notions of consistency, but at prohibitive training-time cost. Our framework is agnostic to the exact similarity function, allowing practitioners to substitute stronger semantic rewards (e.g., BERTScore, an LLM critic) depending on their compute budget.

\begin{table}[t]
\centering
\caption{\small
\textbf{Effect of a semantic similarity reward on Con-RAG: BLEU vs.\ BERTScore} (NQ, LLaMA-3.1-8B).
We replace BLEU with BERTScore as the similarity function in the group similarity reward. BERTScore achieves comparable LLM-Judge consistency and accuracy, but with lower lexical consistency, since it permits semantically equivalent but lexically diverse outputs. This comes at a substantially higher compute cost, as each comparison requires a BERT encoder forward pass. CPU and GPU power are logged using CodeCarbon~\cite{benoit_courty_2024_11171501}.}
\small
\begin{adjustbox}{max width=\textwidth}
\begin{tabular}{l|cc|cc|cc|cc}
\toprule
\multirow{2}{*}{\textbf{Reward Metric}}
& \multicolumn{2}{c}{\textbf{Accuracy (\%)}}
& \multicolumn{2}{c}{\textbf{End-to-End Cons. (\%)}}
& \multicolumn{2}{c}{\textbf{Generator Cons. (\%)}}
& \multicolumn{2}{c}{\textbf{Compute}} \\
\cmidrule(lr){2-3}\cmidrule(lr){4-5}\cmidrule(lr){6-7}\cmidrule(lr){8-9}
& ROUGE & LLM-Acc & Lexical & Inform. & Lexical & Inform. & CPU (W) & GPU (W) \\
\midrule
BLEU (default) & 23.5 & 59.0 & \textbf{22.8} & 59.3 & \textbf{32.3} & \textbf{75.5} & 121 & 459 \\
BERTScore      & \textbf{23.8} & \textbf{60.0} & 18.5 & \textbf{60.1} & 27.4 & 74.8 & 141 & 629 \\
\bottomrule
\end{tabular}
\end{adjustbox}
\label{tab:bertscore-reward}
\end{table}

\subsection{Why Does Consistency Training Improve Accuracy?}\label{apx:accuracy}
Our goal is to improve information consistency without compromising accuracy; accuracy gains, particularly those obtained without ground-truth supervision, are a welcome byproduct rather than the primary objective. These gains are not simply a consequence of reduced output diversity. We hypothesize they arise from two complementary mechanisms.

\textit{(1) Implicit data augmentation.} By tying multiple paraphrased inputs of the same query together during training, our approach acts as a form of regularization: the objective pushes the model to capture the invariant semantic structure across paraphrases rather than their lexical differences. This is analogous to data augmentation, where label-preserving transformations such as cropping and flipping in vision~\citep{shorten2019survey}, noise injection in speech~\citep{ko2015audio}, and paraphrasing in NLP~\citep{feng2021survey} improve generalization by exposing the model to diverse realizations of the same underlying input. As a result, the model becomes more robust to paraphrasing.

\textit{(2) Consistency as a regularizer against hallucination.} Correct answers are inherently more reproducible across paraphrases than hallucinations: a hallucinated answer is unlikely to appear consistently across different phrasings and retrieved documents. This premise underlies self-consistency decoding~\citep{wang2023selfconsistency}, semantic-entropy hallucination detection~\citep{kuhn2023semantic}, and majority-vote rewards~\citep{zuo2026ttrl}. Rewarding cross-paraphrase agreement therefore favors content grounded in evidence shared across retrievals and filters out noise, similar to a ``wisdom of crowds'' effect. The KL term prevents the model from collapsing to trivial but consistent outputs.

Empirically, consistency-only training ($\alpha{=}1,\gamma{=}0$) already improves accuracy over RAG on TriviaQA (EM 51.5 vs.\ 42.0; Table~\ref{tab:abl-acc}), and on ELI5, Con-RAG trained without any ground truth achieves the highest LLM-judged accuracy of all methods (78.0 vs.\ 74.0 for RAG; Table~\ref{tab:main-results-longform}). On NQ, accuracy is preserved within two points of RAG (59.0 vs.\ 61.0) alongside large consistency gains.

\textit{Consistent but wrong outputs.} Our method can lead to consistent but wrong outputs on certain inputs. Even in such cases, consistency is preferable to being both inconsistent and wrong, since the latter produces more unpredictable and contradictory outputs. In our experiments, we do not observe this trade-off dominating in practice: consistency improvements are accompanied by improved or preserved accuracy relative to RAG. Future work could more deeply characterize when consistency and accuracy align or diverge, especially in long-form, open-ended settings.

\subsection{Out-of-Distribution Evaluations}\label{apx:ood}
A natural concern with a consistency-based objective is that the policy may improve group similarity rewards by locking onto a set of generic, dataset-specific safe outputs rather than learning a transferable consistency behavior. To test this, we conduct out-of-distribution (OOD) evaluations on long-form QA: we train Con-RAG on one long-form dataset and evaluate it on the other (ELI5$\rightarrow$NQ and NQ$\rightarrow$ELI5). If Con-RAG had collapsed onto dataset-specific outputs, these outputs would carry over to the new distribution, where they would no longer be correct, and accuracy would fall sharply while consistency stayed artificially high.

We do not observe this behavior (see Table~\ref{tab:ood}). On ELI5, the model trained on NQ matches the RAG baseline in LLM-judged accuracy while improving information consistency at both the end-to-end ($62.8 \rightarrow 69.5$) and generator ($74.2 \rightarrow 77.6$) levels. On NQ, the model trained on ELI5 remains close to the RAG baseline in accuracy ($57.0$ vs. $61.0$) while improving information consistency from $53.2 \rightarrow 56.8$ (end-to-end) and $68.2 \rightarrow 72.1$ (generator). Lexical consistency improves over RAG in both directions as well. As expected, in-distribution training yields the largest gains, but the transferred models still improve over the baseline on every consistency metric. This indicates that Con-RAG learns a transferable consistency behavior rather than concentrating on dataset-specific outputs. Training on a mixed-task paraphrase corpus to obtain a single broadly consistent model is a natural extension, and would further reduce the need for per-dataset training.

\begin{table}[t]
\centering
\caption{\small
\textbf{Out-of-distribution generalization of Con-RAG on long-form QA} (LLaMA-3.1-8B).
Con-RAG is trained on the dataset in the \emph{Train} column and evaluated on the dataset in the \emph{Eval} column (ID: in-distribution; OOD: out-of-distribution). Transferred models preserve accuracy close to the RAG baseline on the unseen dataset while still improving both end-to-end and generator consistency, indicating a transferable consistency behavior rather than dataset-specific output concentration.}
\small
\setlength{\tabcolsep}{4pt}
\begin{adjustbox}{max width=\textwidth}
\begin{tabular}{lll|cc|cc|cc}
\toprule
 & & & \multicolumn{2}{c}{\textbf{Accuracy (\%)}}
 & \multicolumn{2}{c}{\shortstack[c]{\textbf{End-to-End}\\\textbf{Consistency (\%)}}}
 & \multicolumn{2}{c}{\shortstack[c]{\textbf{Generator (LLM)}\\\textbf{Consistency (\%)}}} \\
\cmidrule(lr){4-5} \cmidrule(lr){6-7} \cmidrule(lr){8-9}
\textbf{Eval} & \textbf{Method} & \textbf{Train}
& ROUGE & LLM-Acc
& Lexical & Inform.
& Lexical & Inform. \\
\midrule
\multirow{3}{*}{ELI5}
& RAG     & --          & 21.9 & 74.0 & 8.6  & 62.8 & 15.1 & 74.2 \\
& Con-RAG & ELI5 (ID)   & \textbf{24.2} & \textbf{78.0} & \textbf{14.6} & \textbf{72.7} & \textbf{21.7} & \textbf{80.8} \\
& Con-RAG & NQ (OOD)    & 23.1 & 74.0 & 12.8 & 69.5 & 19.4 & 77.6 \\
\midrule
\multirow{3}{*}{NQ}
& RAG     & --          & \textbf{26.9} & \textbf{61.0} & 14.2 & 53.2 & 22.4 & 68.2 \\
& Con-RAG & NQ (ID)     & 23.5 & 59.0 & \textbf{22.8} & \textbf{59.3} & \textbf{32.3} & \textbf{75.5} \\
& Con-RAG & ELI5 (OOD)  & 22.8 & 57.0 & 19.5 & 56.8 & 28.9 & 72.1 \\
\bottomrule
\end{tabular}
\end{adjustbox}
\label{tab:ood}
\end{table}

\subsection{Choosing the KL Coefficient Hyperparameter $\beta$}\label{apx:kl}
A consistency-only objective can in principle be maximized by converging to answers that agree across paraphrases but are trivial or wrong; a KL penalty to the reference policy mitigates this risk. When rewards are verifiable against ground truth, recent GRPO work has progressively weakened this penalty, from $\beta{=}0.04$~\cite{shao2024deepseekmath} to $\beta{=}0.001$~\cite{guo2025deepseekr1}, and in several cases removed it entirely~\cite{yu2026dapo,hu2025open,liu2025understanding}, since a rule-based reward offers no learned model for the policy to exploit. Our label-free setting lacks this safeguard. Below, we summarize the evidence from our experiments and give practical guidelines for setting $\beta$.

\textit{When an accuracy term is present, $\beta{=}0$ suffices.} In short-form and multi-hop QA, the token-F1 reward anchors outputs to correct content, and we train without KL, following~\cite{hu2025open}. Even without the accuracy term, consistency-only training on TriviaQA does not collapse: it improves EM over RAG (51.5 vs.\ 42.0; Table~\ref{tab:abl-acc}). Short answers leave little room for generic outputs, and the paraphrase-level reward still favors the answer that is reproducible across retrieved contexts.

\textit{Without ground truth, a small $\beta$ prevents degenerate consensus.} For long-form QA we use $\beta{=}0.05$. Our metrics detect collapse when it occurs: its signature is rising lexical consistency with falling information consistency and accuracy. SFT shows exactly this pattern on NQ (lexical 37.1, LLM-Judge 36.2, LLM-Acc 22.0; Table~\ref{tab:main-results-longform}). Con-RAG with $\beta{=}0.05$ shows the opposite pattern, with LLM-Judge consistency well above lexical consistency (59.3 vs.\ 22.8) and accuracy preserved (59.0 vs.\ 61.0 for RAG); on ELI5 it achieves the highest LLM-Acc of any method (78.0). The setting is also robust to other design choices: across the $(\kappa, s)$ sweep, LLM-Acc stays within 59.0--63.0 (Table~\ref{tab:nq-kappa-s}), and replacing BLEU with BERTScore leaves accuracy unchanged (Appendix~\ref{apx:bertscore}).

\textit{Guidelines.}
(i) If gold answers are available, include the accuracy term and set $\beta{=}0$.
(ii) Without gold answers, start from a small $\beta$ (we use 0.05) and select it on a small held-out set scored by an LLM judge for accuracy, not by training reward, since the consistency reward increases monotonically as $\beta$ decreases.
(iii) During training, monitor collapse signatures: lexical consistency rising while LLM-Judge consistency or judged accuracy falls, shrinking response length or distinct-$n$, or rapidly growing KL to the reference. If any appears, increase $\beta$.
 
\subsection{Robustness to Retrieval Depth and Decoding Strategy}\label{apx:robustness_pipeline}

\textbf{Sensitivity to the number of retrieved documents (top-$k$).}
Con-RAG is designed to be retriever-agnostic. Our RAG pipeline (e5-base-v2 embeddings, top-5 retrieval, 512-token chunks, KILT Wikipedia corpus) was configured to first maximize accuracy over original queries before conducting any consistency analysis on paraphrased queries, mirroring how a RAG system would be set up in practice. This setup is also consistent with related work on RAG evaluation and optimization~\cite{yue2024inference}. To evaluate the sensitivity of Con-RAG to this pipeline parameter, we vary the number of retrieved documents $k \in \{5, 10, 15\}$ on NQ (LLaMA-3.1-8B). Results are summarized in Table~\ref{tab:topk-sensitivity}. Accuracy slightly improves with larger $k$ (more relevant evidence), while consistency shows a mild decrease, as additional noisy and tangential content increases the divergence surface across paraphrased queries. Importantly, Con-RAG maintains large consistency gains over the RAG baseline across all $k$ values (e.g., E2E LLM-Judge consistency of $56.1$--$59.3$ vs.\ $53.2$ for RAG), confirming robustness to the pipeline configuration. Since our method is agnostic to the retriever, we do not expect results to change substantially with different retrieval mechanisms.

\begin{table}[t]
\centering
\caption{\small
\textbf{Sensitivity of Con-RAG to the number of retrieved documents (top-$k$)} (NQ, LLaMA-3.1-8B).
Accuracy slightly improves with larger $k$, while consistency mildly decreases as more tangential content is introduced. Con-RAG maintains large consistency gains over the RAG baseline across all $k$ values.}
\small
\begin{tabular}{l|cc|cc|cc}
\toprule
\multirow{2}{*}{\textbf{Setting}}
& \multicolumn{2}{c}{\textbf{Accuracy (\%)}}
& \multicolumn{2}{c}{\textbf{\shortstack[c]{\textbf{End-to-End}\\\textbf{Consistency (\%)}}}}
& \multicolumn{2}{c}{\shortstack[c]{\textbf{Generator (LLM)}\\\textbf{Consistency (\%)}}} \\
\cmidrule(lr){2-3}\cmidrule(lr){4-5}\cmidrule(lr){6-7}
& ROUGE & LLM-Acc & Lexical & Inform. & Lexical & Inform. \\
\midrule
RAG ($k{=}5$)              & \textbf{26.9} & \textbf{61.0} & 14.2 & 53.2 & 22.4 & 68.2 \\
\midrule
Con-RAG ($k{=}5$, default) & 23.5 & 59.0 & \textbf{22.8} & \textbf{59.3} & \textbf{32.3} & \textbf{75.5} \\
Con-RAG ($k{=}10$)         & 24.1 & 60.0 & 21.3 & 57.8 & 31.0 & 74.2 \\
Con-RAG ($k{=}15$)         & 24.4 & \textbf{61.0} & 19.7 & 56.1 & 29.5 & 72.8 \\
\bottomrule
\end{tabular}
\label{tab:topk-sensitivity}
\end{table}

\textbf{Effect of decoding strategy on Con-RAG.}
In our main experiments, we use deterministic decoding at inference specifically to isolate the inconsistency due to paraphrasing from sampling stochasticity. This is a deliberate methodological choice and is standard practice in consistency evaluation~\cite{elazar-etal-2021-measuring,raj2025improving}. We note that during training, we use non-deterministic decoding to generate diverse rollouts; deterministic decoding is only used at inference. To evaluate practicality under more common decoding strategies, we compare greedy decoding against top-$p$ sampling ($p{=}0.9$), top-$k$ sampling ($k{=}50$), and temperature sampling ($T{=}0.5$) for Con-RAG on NQ (LLaMA-3.1-8B) (see Table~\ref{tab:decoding-conrag}). Stochastic decoding introduces a moderate drop in lexical consistency, as expected, since sampling adds variability on top of paraphrase-induced variation. However, LLM-Judge consistency and accuracy remain stable, and Con-RAG remains above the greedy RAG baseline (E2E LLM-Judge $53.2$, Gen.\ LLM-Judge $68.2$) under every decoding strategy. Temperature sampling at $T{=}0.5$ is closest to greedy decoding and even slightly improves LLM-Acc ($60.0$ vs.\ $59.0$).

We further evaluate how inference-time stochasticity affects standard RAG by sweeping temperature values $T \in \{0.0, 0.5, 1.0, 2.0\}$ on ELI5 (Qwen-2.5-3B) (see Table~\ref{tab:abl-temp}). Moderate temperature ($T{=}0.5$) improves LLM-Judge and lexical consistency compared to deterministic decoding while preserving accuracy, whereas higher temperatures ($T \geq 1.0$) degrade both consistency and accuracy, with outputs at $T{=}2.0$ nearly collapsing.

\begin{table}[h]
\centering
\caption{\small
\textbf{Effect of decoding strategy on Con-RAG} (NQ, LLaMA-3.1-8B).
Stochastic decoding moderately reduces lexical consistency, while LLM-Judge consistency and accuracy remain stable. Con-RAG remains above the greedy RAG baseline (E2E LLM-Judge 53.2, Gen.\ LLM-Judge 68.2) under all decoding strategies.}
\small
\begin{tabular}{l|cc|cc|cc}
\toprule
\multirow{2}{*}{\textbf{Decoding}}
& \multicolumn{2}{c}{\textbf{Accuracy (\%)}}
& \multicolumn{2}{c}{\textbf{End-to-End Cons. (\%)}}
& \multicolumn{2}{c}{\textbf{Generator Cons. (\%)}} \\
\cmidrule(lr){2-3}\cmidrule(lr){4-5}\cmidrule(lr){6-7}
& ROUGE & LLM-Acc & Lexical & Inform. & Lexical & Inform. \\
\midrule
Greedy ($T{=}0.0$, default) & \textbf{23.5} & 59.0 & \textbf{22.8} & \textbf{59.3} & \textbf{32.3} & \textbf{75.5} \\
Top-$p$ ($p{=}0.9$)         & 22.8 & 58.0 & 18.4 & 56.8 & 28.1 & 72.3 \\
Top-$k$ ($k{=}50$)          & 22.6 & 57.0 & 17.9 & 55.5 & 27.4 & 71.8 \\
Temperature ($T{=}0.5$)     & 23.1 & \textbf{60.0} & 20.2 & 58.1 & 30.5 & 74.0 \\
\bottomrule
\end{tabular}
\label{tab:decoding-conrag}
\end{table}

\begin{table}[h]
\centering
\caption{\small
\textbf{Effect of inference temperature on standard RAG} (ELI5, Qwen-2.5-3B).
We vary only the decoding temperature $T$ at inference to study its effect on consistency and accuracy. Moderate temperature ($T{=}0.5$) improves LLM-Judge and lexical consistency compared to deterministic decoding ($T{=}0.0$), while preserving accuracy. However, higher temperatures ($T \geq 1.0$) degrade both consistency and accuracy, with outputs at $T{=}2.0$ nearly collapsing.}
\small
\begin{tabular}{c|cc|cc|cc}
\toprule
\multirow{2}{*}{$T$}
& \multicolumn{2}{c}{\textbf{Accuracy (\%)}}
& \multicolumn{2}{c}{\textbf{End-to-End Cons. (\%)}}
& \multicolumn{2}{c}{\textbf{Generator Cons. (\%)}} \\
\cmidrule(lr){2-3}\cmidrule(lr){4-5}\cmidrule(lr){6-7}
& ROUGE & LLM-Acc & Lexical & Inform. & Lexical & Inform. \\
\midrule
0.0 & \textbf{22.1} & 38.0 & 6.6 & 35.3 & 14.4 & 62.3 \\
0.5 & 21.4 & \textbf{52.0} & \textbf{10.4} & \textbf{37.7} & \textbf{15.2} & \textbf{65.3} \\
1.0 & 21.8 & 48.0 & 2.5 & 34.0 & 5.2 & 59.5 \\
2.0 & 6.1  & 0.0  & 0.1 & 2.0  & 0.2 & 1.5 \\
\bottomrule
\end{tabular}
\label{tab:abl-temp}
\end{table}

\subsection{Training Cost under Distribution Drift}\label{apx:drift}
A practical concern is the cost of retraining Con-RAG when the query distribution shifts rapidly. We have not studied retraining cadence under drift directly, but several factors reduce the practical cost.

\textit{(i) Retraining requires no annotation.} Under drift, the dominant recurring cost for supervised alternatives (SFT, CoG, CAT) is producing reference answers for the new distribution. PS-GRPO's group similarity reward is computed from mutual agreement among rollouts across paraphrases, so a refresh needs only new queries and their paraphrases.

\textit{(ii) Retraining may not be needed for every shift.} Our out-of-distribution results (Table~\ref{tab:ood}) suggest the learned behavior is not distribution-specific: a model trained on NQ and evaluated on ELI5 retains much of its consistency gain over RAG at comparable accuracy. Training on a mixed-task paraphrase corpus to obtain a single, broadly consistent model is a natural extension that would further reduce retraining frequency.

\textit{(iii) A refresh can run at the low-cost end of the compute frontier.} The subsampling ablation (Table~\ref{tab:nq-kappa-s}) shows the minimal setting is already competitive: $(\kappa{=}1, s{=}1)$ achieves 60.0 LLM-Acc and 70.8 generator LLM-Judge consistency at 441\,W GPU power, compared with 63.0 / 76.2 at 492\,W for $(\kappa{=}5, s{=}5)$.
Distribution drift is a general challenge for every trained component of a RAG pipeline, and retrieval quality and accuracy are likely to degrade before consistency becomes the binding concern. Con-RAG is therefore not uniquely exposed, and its label-free refresh makes it among the cheaper components to keep current.

\newpage
\section*{NeurIPS Paper Checklist}

\begin{enumerate}

\item {\bf Claims}
    \item[] Question: Do the main claims made in the abstract and introduction accurately reflect the paper's contributions and scope?
    \item[] Answer: \answerYes{} 
    \item[] Justification: Abstract and introduction match papers claims. Details in Main Section \ref{sec:method} 

\item {\bf Limitations}
    \item[] Question: Does the paper discuss the limitations of the work performed by the authors?
    \item[] Answer: \answerYes{} 
    \item[] Justification: Discussed in Section \ref{sec:disc}.

\item {\bf Theory assumptions and proofs}
    \item[] Question: For each theoretical result, does the paper provide the full set of assumptions and a complete (and correct) proof?
    \item[] Answer: \answerNA{} 
    \item[] Justification: The paper has no theoretical results.

    \item {\bf Experimental result reproducibility}
    \item[] Question: Does the paper fully disclose all the information needed to reproduce the main experimental results of the paper to the extent that it affects the main claims and/or conclusions of the paper (regardless of whether the code and data are provided or not)?
    \item[] Answer: \answerYes{} 
    \item[] Justification: Experiment details are in Section~\ref{sec:exp}.

\item {\bf Open access to data and code}
    \item[] Question: Does the paper provide open access to the data and code, with sufficient instructions to faithfully reproduce the main experimental results, as described in supplemental material?
    \item[] Answer: \answerYes{} 
\item[] Justification: Code for paraphrase generation, PS-GRPO training, and evaluation is released at \url{https://github.com/CapitalOne-Research/con-rag}. All datasets used are public, and prompts and hyperparameters are given in Section~\ref{sec:exp} and the Appendix.


\item {\bf Experimental setting/details}
    \item[] Question: Does the paper specify all the training and test details (e.g., data splits, hyperparameters, how they were chosen, type of optimizer) necessary to understand the results?
    \item[] Answer: \answerYes{} 
    \item[] Justification: Experimental details are in Appendix Section~\ref{apx:com_PSGRPO_GRPO}

\item {\bf Experiment statistical significance}
    \item[] Question: Does the paper report error bars suitably and correctly defined or other appropriate information about the statistical significance of the experiments?
    \item[] Answer: \answerNA{} 
    \item[] Justification: Does not apply to our experimental setting.

\item {\bf Experiments compute resources}
    \item[] Question: For each experiment, does the paper provide sufficient information on the computer resources (type of compute workers, memory, time of execution) needed to reproduce the experiments?
    \item[] Answer: \answerYes{} 
    \item[] Justification: Details in Experimental section and Appendix.
    
\item {\bf Code of ethics}
    \item[] Question: Does the research conducted in the paper conform, in every respect, with the NeurIPS Code of Ethics \url{https://neurips.cc/public/EthicsGuidelines}?
    \item[] Answer: \answerYes{} 
    \item[] Justification: 
    \item[] Guidelines: Conforms with the NeurIPS Code of Ethics to the best of our knowledge.
    \begin{itemize}
        \item The answer \answerNA{} means that the authors have not reviewed the NeurIPS Code of Ethics.
        \item If the authors answer \answerNo, they should explain the special circumstances that require a deviation from the Code of Ethics.
        \item The authors should make sure to preserve anonymity (e.g., if there is a special consideration due to laws or regulations in their jurisdiction).
    \end{itemize}

\item {\bf Broader impacts}
    \item[] Question: Does the paper discuss both potential positive societal impacts and negative societal impacts of the work performed?
    \item[] Answer: \answerYes{} 
    \item[] Justification: In Discussion section and Impact Statement.
    
\item {\bf Safeguards}
    \item[] Question: Does the paper describe safeguards that have been put in place for responsible release of data or models that have a high risk for misuse (e.g., pre-trained language models, image generators, or scraped datasets)?
    \item[] Answer: \answerNA{} 
    \item[] Justification: Not applicable to this work.

\item {\bf Licenses for existing assets}
    \item[] Question: Are the creators or original owners of assets (e.g., code, data, models), used in the paper, properly credited and are the license and terms of use explicitly mentioned and properly respected?
    \item[] Answer: \answerYes{} 
    \item[] Justification: Assets are properly cited. 

\item {\bf New assets}
    \item[] Question: Are new assets introduced in the paper well documented and is the documentation provided alongside the assets?
    \item[] Answer: \answerNA{} 
    \item[] Justification: Paper does not release new assets.

\item {\bf Crowdsourcing and research with human subjects}
    \item[] Question: For crowdsourcing experiments and research with human subjects, does the paper include the full text of instructions given to participants and screenshots, if applicable, as well as details about compensation (if any)? 
    \item[] Answer: \answerNA{} 
    \item[] Justification: Not applicable to this work.

\item {\bf Institutional review board (IRB) approvals or equivalent for research with human subjects}
    \item[] Question: Does the paper describe potential risks incurred by study participants, whether such risks were disclosed to the subjects, and whether Institutional Review Board (IRB) approvals (or an equivalent approval/review based on the requirements of your country or institution) were obtained?
    \item[] Answer: \answerNA{} 
    \item[] Justification: Not applicable to this work.

\item {\bf Declaration of LLM usage}
    \item[] Question: Does the paper describe the usage of LLMs if it is an important, original, or non-standard component of the core methods in this research? Note that if the LLM is used only for writing, editing, or formatting purposes and does \emph{not} impact the core methodology, scientific rigor, or originality of the research, declaration is not required.
    \item[] Answer: \answerYes{} 
    \item[] Justification: LLMs used for evaluation.

\end{enumerate}

\end{document}